%% file: neurips_data_2024.tex
\documentclass{article}

\PassOptionsToPackage{numbers, compress}{natbib}

\usepackage[preprint]{neurips_data_2024}

\usepackage[utf8]{inputenc} 
\usepackage[T1]{fontenc}    
\usepackage{hyperref}       
\usepackage{url}            
\usepackage{booktabs}       
\usepackage{amsfonts}       
\usepackage{nicefrac}       
\usepackage{microtype}      
\usepackage{xcolor}         
\usepackage{graphicx}
\usepackage{amsmath}
\usepackage{cleveref}
\crefname{section}{§}{§§}
\Crefname{section}{§}{§§}
\usepackage{array}

\usepackage{multicol}
\usepackage{multirow}
\usepackage{tabularx}
\usepackage{enumitem}
\usepackage{longtable}
\usepackage{graphicx}
\usepackage{array}
\usepackage{amssymb}
\usepackage{multirow}
\usepackage{xcolor,colortbl}
\usepackage{xspace}
\usepackage{makecell}
\usepackage{bbding}
\usepackage{boldline}
\usepackage{pifont}
\usepackage{tabularx}
\usepackage{graphicx}
\usepackage{wrapfig}
\usepackage[export]{adjustbox}
\usepackage{fvextra}
\usepackage[most,breakable]{tcolorbox}
\usepackage[hypcap=false]{caption}
\usepackage{pgfplots}
\usepackage{arydshln}
\usepackage{subcaption}
\pgfplotsset{compat=newest}
\usepgfplotslibrary{units}

\newcommand{\ours}[0]{Spider2-V\xspace}
\newcommand{\tasknum}[0]{$494$\xspace}
\newcommand{\toolnum}[0]{$20$\xspace}

\newcommand{\initnum}[0]{$170$\xspace}
\newcommand{\evalnum}[0]{$151$\xspace}
\newcommand{\tutorialnum}[0]{$217$\xspace}
\newcommand{\docnum}[0]{$11,231$\xspace}
\newcommand{\fulldocnum}[0]{$21,239$\xspace}

\newcommand{\sotamodel}[0]{GPT-4V\xspace}

\newcommand{\verbosesota}[0]{$16.2\%$\xspace}
\newcommand{\sota}[0]{$14.0\%$\xspace}

\title{\ours: How Far Are Multimodal Agents From Automating Data Science and Engineering Workflows?}

\author{
\small
  {\bf
    Ruisheng Cao\thanks{\ \ Work done while interning at the University of Hong Kong.} 
    $^{\hspace{.1em}{\color{purple}\boldsymbol{12}}}$
    \enskip
    Fangyu Lei
    $^{{\color{purple}\boldsymbol{1}}}$
    \enskip
    Haoyuan Wu
    $^{{\color{purple}\boldsymbol{1}}}$
    \enskip
    Jixuan Chen
    $^{{\color{purple}\boldsymbol{1}}}$
    \enskip
    Yeqiao Fu
    $^{{\color{purple}\boldsymbol{1}}}$
    \enskip
    Hongcheng Gao
    $^{{\color{purple}\boldsymbol{1}}}$
    \vspace{4pt}
  } \\
\small
  {
  \bf
    Xinzhuang Xiong
    $^{{\color{purple}\boldsymbol{1}}}$
    \enskip
    Hanchong Zhang
    $^{{\color{purple}\boldsymbol{2}}}$
    \enskip
    Yuchen Mao
    $^{{\color{purple}\boldsymbol{1}}}$
    \enskip
    Wenjing Hu
    $^{{\color{purple}\boldsymbol{1}}}$
    \enskip
    Tianbao Xie
    $^{{\color{purple}\boldsymbol{1}}}$
    Hongshen Xu
    $^{{\color{purple}\boldsymbol{2}}}$
    \vspace{4pt}
  } \\
\small
  {
  \bf
    Danyang Zhang
    $^{{\color{purple}\boldsymbol{12}}}$
    \enskip
    Sida Wang
    \enskip
    Ruoxi Sun
    $^{{\color{purple}\boldsymbol{3}}}$
    \enskip
    Pengcheng Yin
    $^{{\color{purple}\boldsymbol{4}}}$
    \enskip
    Caiming Xiong
    $^{{\color{purple}\boldsymbol{5}}}$
    \enskip
    Ansong Ni
    $^{{\color{purple}\boldsymbol{6}}}$
    \vspace{4pt}
  } \\
\small
  {
  \bf
    Qian Liu
    $^{{\color{purple}\boldsymbol{7}}}$
    \enskip
    Victor Zhong
    $^{{\color{purple}\boldsymbol{8}}}$
    \enskip
    Lu Chen
    $^{{\color{purple}\boldsymbol{2}}}$
    \enskip
    Kai Yu
    $^{{\color{purple}\boldsymbol{2}}}$
    \enskip
    Tao Yu
    $^{{\color{purple}\boldsymbol{1}}}$
    \vspace{4pt}
  } \\
\small
  {
    \hspace{-12pt}
    $^{\color{purple}\boldsymbol{1}}$ The University of Hong Kong \quad 
    $^{\color{purple}\boldsymbol{2}}$ Shanghai Jiao Tong University
    \vspace{2pt}
    }\\
\small
    {
    $^{\color{purple}\boldsymbol{3}}$ Google Cloud AI Research
    \quad
    $^{\color{purple}\boldsymbol{4}}$ Google DeepMind
    \quad
    $^{\color{purple}\boldsymbol{5}}$ Salesforce Research
    \vspace{2pt}
  } \\
\small
  {
    $^{\color{purple}\boldsymbol{6}}$ Yale University
    \vspace{2pt}
    \quad
    $^{\color{purple}\boldsymbol{7}}$ Sea AI Lab
    \quad
    $^{\color{purple}\boldsymbol{8}}$ University of Waterloo
  } \\
}

\begin{document}

\maketitle

\input{0.abstract_new}

\input{1.introduction_new}

\input{2.environment}

\input{3.benchmark}
\input{4.experiments}

\input{5.related_work}

\input{6.conclusion}



\bibliographystyle{plainnat}
\bibliography{reference}

\clearpage
\appendix
\section{Relevant URLs}
\begin{figure}[htbp]
    \centering
    \includegraphics[width=0.95\textwidth]{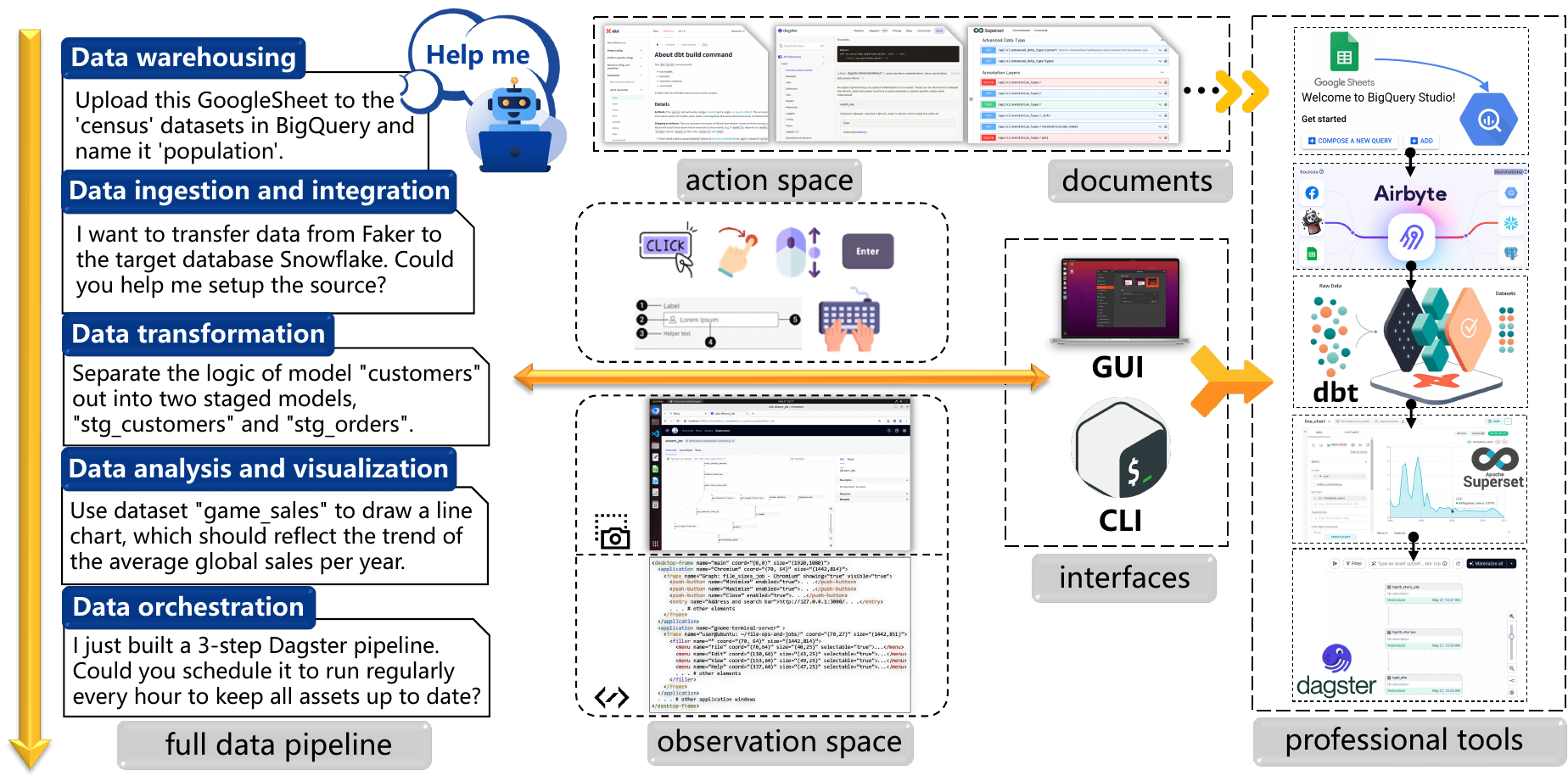}
    \caption{Overview of \ours, which includes task examples across the full data pipeline, an executable computer environment, and a document warehouse for agent retrieval.}
    \label{fig:app_overview}
\end{figure}
\paragraph{Github Repository} The task examples, environment, documents, code and experiments are publicly available in Github repository \url{https://github.com/xlang-ai/Spider2-V} under {\bf Apache-2.0 license}. Both the environment and task examples will be maintained by the authors continuously.

Concretely, the environment code is adapted from previous work \textsc{OSWorld}~\citep{osworld}, which is released under Apache-2.0 license. A non-exhaustive list of artifacts~(or task examples) used in \ours includes: 1) SheetCopilot~\citep{sheetcopilot} which is released under GPL-3.0 license, 2) WorkArena~\citep{workarena} which is distributed under Apache-2.0 license, and 3) official tutorials or guides on professional applications~(e.g., {\tt dbt}, {\tt Airflow}, {\tt Dagster}, {\tt Superset}, etc.). These tutorials are free to use and publicly available. For those enterprise applications which require real accounts, namely {\tt BigQuery}, {\tt Snowflake}, {\tt dbt-cloud} and {\tt ServiceNow}, we only exploit their sandbox functions or free-trials without introducing any extra cost or privacy issues.

\paragraph{Project Website} We also build a project website \url{https://spider2-v.github.io/} based on Nerfies~\citep{nerfies} template which is free-to-use and licensed under a Creative Commons Attribution-ShareAlike 4.0 International License. On this website, we provide a high-level overview of \ours, the leaderboard of the benchmark and more concrete dynamic task demonstrations.

The authors declare that the benchmark collection and usage strictly obey the aforementioned licenses.
\input{appendices/softwares}
\input{appendices/documents}
\input{appendices/environment}
\input{appendices/task_example}
\input{appendices/examples}
\input{appendices/prompts}

\end{document}

%% file: 0.abstract_new.tex
\vspace{-1em}
\begin{abstract}

Data science and engineering workflows often span multiple stages, from warehousing to orchestration, using tools like {\tt BigQuery}, {\tt dbt}, and {\tt Airbyte}. As vision language models (VLMs) advance in multimodal understanding and code generation, VLM-based agents could potentially automate these workflows by generating SQL queries, Python code, and GUI operations. This automation can improve the productivity of experts while democratizing access to large-scale data analysis.
In this paper, we introduce \ours, the first multimodal agent benchmark focusing on professional data science and engineering workflows, featuring \tasknum real-world tasks in authentic computer environments and incorporating \toolnum enterprise-level professional applications.
These tasks, derived from real-world use cases, evaluate the ability of a multimodal agent to perform data-related tasks by writing code and managing the GUI in enterprise data software systems.
To balance realistic simulation with evaluation simplicity, we devote significant effort to developing automatic configurations for task setup and carefully crafting evaluation metrics for each task.
Furthermore, we supplement multimodal agents with comprehensive documents of these enterprise data software systems.
Our empirical evaluation reveals that existing state-of-the-art LLM/VLM-based agents do not reliably automate full data workflows (\sota success). 
Even with step-by-step guidance, these agents still underperform in tasks that require fine-grained, knowledge-intensive GUI actions~($16.2\%$) and involve remote cloud-hosted workspaces~($10.6\%$).
We hope that \ours paves the way for autonomous multimodal agents to transform the automation of data science and engineering workflow.
Our code and data are available at \url{https://spider2-v.github.io}.

\end{abstract}

%% file: 1.introduction_new.tex
\section{Introduction}

\begin{figure}[htbp]
    \centering
    \includegraphics[width=0.98\textwidth]{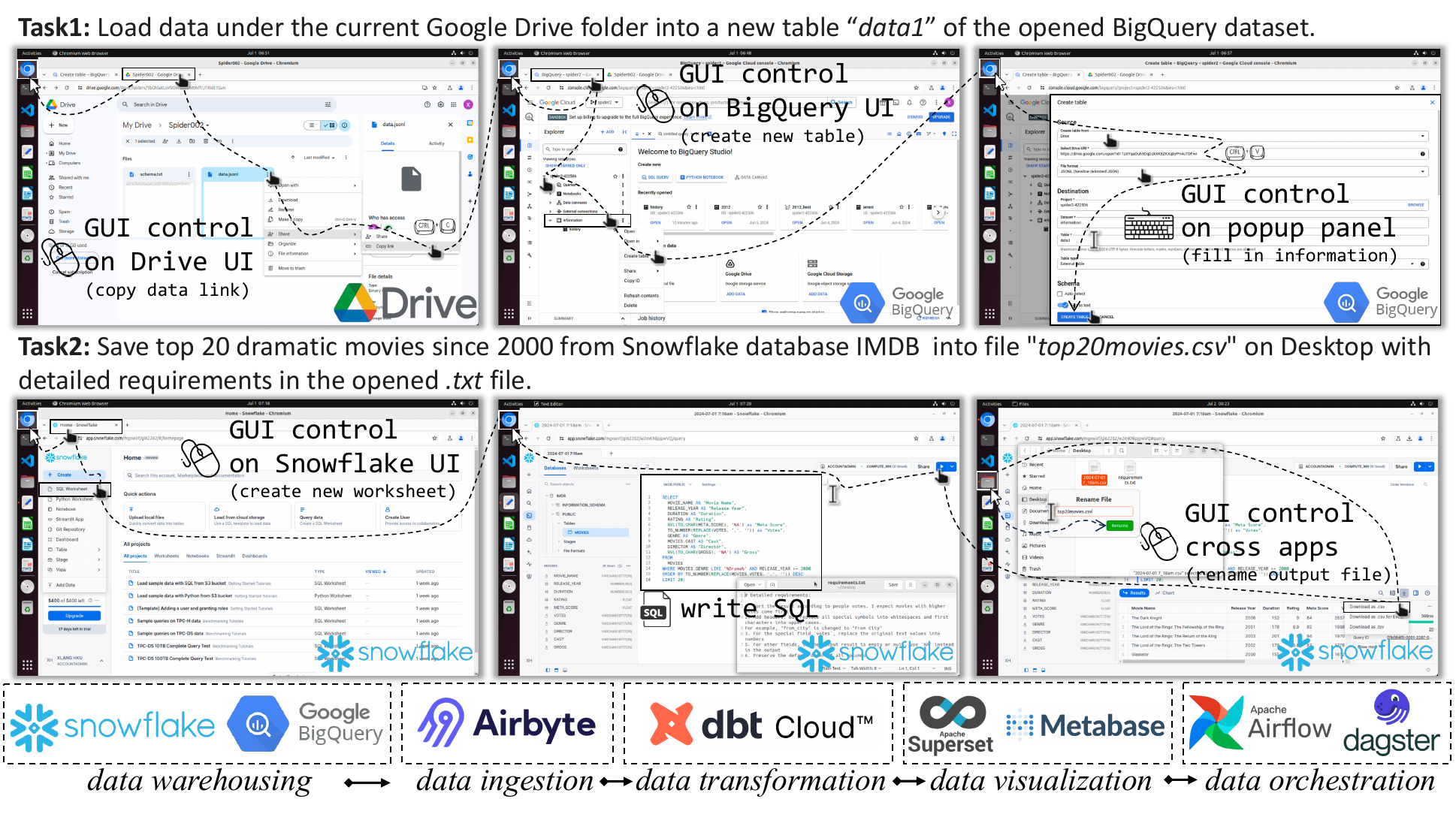}
    \caption{\ours is a multimodal agent benchmark spanning across complete data science and engineering workflows~(\emph{e.g.}, two task examples in the Figure above). It involves various professional enterprise-level applications and includes intensive GUI controls apart from code writing throughout the real-time multi-turn interaction with an executable computer environment.
    }
    \label{fig:intro}
\end{figure}

Data science and engineering pipelines usually rely on professional data software systems such as {\tt BigQuery}, {\tt dbt}, and {\tt Airbyte} to acquire, process, and orchestrate large-scale data.
Utilizing these enterprise systems involves writing SQL and Python code, as well as frequent and repetitive graphical user interface~(GUI) controls, which can be complex even for experienced data scientists and engineers.
With rapid advances in large language models~(LLMs) and vision language models~(VLMs), LLM/VLM-based autonomous agents have the potential to automate these workflows~\cite{yang2024swe,wu2024copilot}, enhancing productivity for data scientists and engineers~\cite{intercode,sheetcopilot} while democratizing access to large-scale data~\cite{ds1000,arcade}.

Previous studies on data agents focused mainly on daily life data processing and analysis by generating code or API calls~\citep{yu2018spider,dabench,sheetagent}, neglecting other crucial stages of data science and engineering (\textit{e.g.,} data ingestion and integration) using enterprise applications (\textit{e.g.,} {\tt Snowflake}, {\tt Airflow}, and {\tt Dagster}). 
Additionally, to complete data workflows, data scientists and engineers often need to navigate multiple professional data systems, combining code writing with intensive GUI controls, such as navigating web pages and clicking buttons~\citep{deng2023mind2web, zhou2023webarena}. 
However, there is currently no benchmark that integrates both code generation and GUI controls for professional data science and engineering.

To address this gap, we propose \ours, the first multimodal agent benchmark covering the entire data science and engineering workflow, involving $494$ real-world tasks in a real-time executable computer environment and \toolnum professional enterprise data software.
\ours aims to evaluate a multimodal agent's ability to perform professional data-related tasks by writing code and managing the GUI in enterprise data software systems, including data warehousing~(\textit{e.g.,} {\tt BigQuery}), data ingestion and integration~(\textit{e.g.,} {\tt Airbyte}), data transformation~(\textit{e.g.,} {\tt dbt}), data analysis and visualization~(\textit{e.g.,} {\tt Superset}), and data orchestration~(\textit{e.g.,} {\tt Dagster}). 
These tasks are derived from real-world practices, such as official tutorials on professional applications and open-source data engineering projects~(with two task examples presented in Figure~\ref{fig:intro}). 
We also supplement retrieval-augmented agents with official documentation and tutorials of these software systems to assess their capability to generalize and learn from these resources.

Each task in \ours is defined within an executable computer environment based on \textsc{OSWorld}~\citep{osworld}, which allows multimodal agents to simulate human actions~(\emph{e.g.}, typing code or clicking buttons) in a realistic setting.
Specifically, a multimodal agent can observe real-time image-style screenshots and text-style accessibility tree of professional data applications in the current workflow and execute its predicted actions in dynamic multi-round interaction with the computer.
This environment is connected to the real-world Internet, allowing the inclusion of professional software requiring authentic user accounts~(\textit{e.g.,} {\tt Snowflake}). 
To ensure reproducible and reliable experiments with this enterprise data software, $10$ authors with computer science backgrounds developed \initnum automatic task setup configurations and \evalnum customized evaluation metrics in total.

We experiment with state-of-the-art LLMs and VLMs including closed-source ones GPT-4 series~\citep{openai2023gpt}, Gemini-Pro-1.5~\cite{reid2024gemini}, Claude-3-Opus~\cite{claude3}, QWen-Max~\cite{qwen} and open-source representatives Mixtral-8x7B~\citep{jiang2024mixtral} and Llama-3-70B~\citep{meta2024llama3}. Performances reveal that even the top-tier VLM~(GPT-4V~\citep{achiam2023gpt}) achieves only \sota success rate.
In the most challenging subset, with action steps exceeding $15$, the performance drops to $1.2\%$. And for those open-source LLMs, the success rate is less than $2\%$. This indicates that existing LLMs or VLMs are still far away from achieving full data workflow automation.
Even provided with an oracle step-by-step plan, the overall performance only increases to \verbosesota. This observation uncovers the poor capability of action grounding~(\emph{e.g.}, identifying the precise coordinates of elements in the current focused application window) for multimodal agents.
Furthermore, extensive analysis~(\cref{sec:analysis}) on \ours demonstrate that these strategies remarkably promote the final performance, which include enhancing the alignment between different observation modalities, introducing feedback on action execution, integrating retrieved document context and enlarging the history trajectory length. These findings lay the groundwork for developing practical multimodal agents that can revolutionize the automation of data science and engineering workflows.

%% file: 2.environment.tex
\section{Executable Computer Environment of \ours}
In this section, we introduce the real-time executable computer environment of \ours, which is built upon virtual machines~(VMs) and adapted from \textsc{OSWorld}~\citep{osworld}. 

\subsection{Task Definition}
Generally, an autonomous data agent is modeled as a partially observable Markov decision process~(POMDP).
Given the current observation $o_{t}\in\mathcal{O}$ which includes a natural language instruction and a screenshot, accessibility tree~({\tt a11ytree}), or their combination, an agent generates an executable action $a_t \in \mathcal{A}$. This action can be clicking on a certain pixel of the screen~(\texttt{CLICK(560, 200)}), or writing code through keyboard~(\texttt{TYPE("ls -lh")}). The execution of $a_{t}$ results in a new state $s_{t+1} \in \mathcal{S}$ (\textit{e.g.}, the updated computer state) and a new partial observation $o_{t+1} \in \mathcal{O}$.
The {\tt a11ytree} is a text-style representation of the desktop environment, which describes the status, position, and text content of each element~(e.g., windows, buttons, and input boxes).
The interaction loop repeats until an action that marks termination (\verb|DONE| or \verb|FAIL|) is generated or the agent reaches the max number of steps. 
See App.~\ref{app:environment} for more details about the observation space and action space.

\subsection{Environment Setup}
\label{sec:env_setup}
\begin{figure}[htbp]
    \centering
    \includegraphics[width=0.9\textwidth]{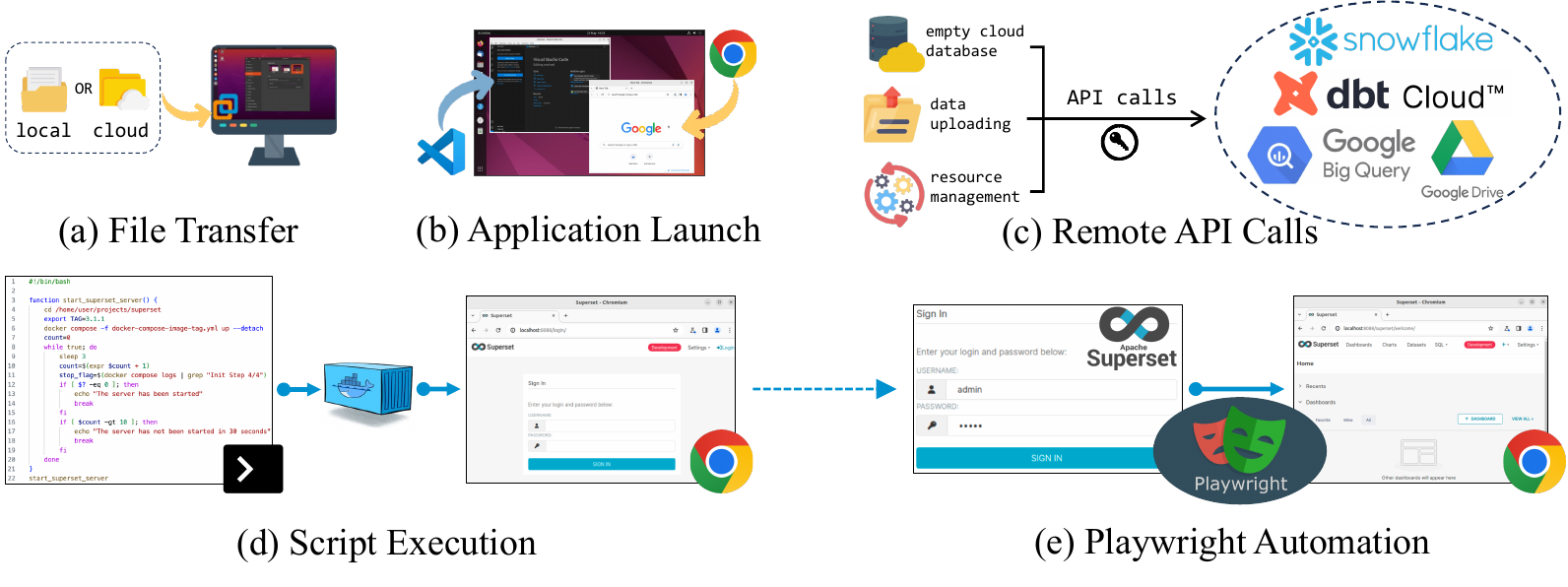}
    \caption{Five common operations to reset the initial environment.}
    \label{fig:env_setup}
\end{figure}
To ensure that an agent starts from a consistent initial state, we invoke a series of function calls based on a pre-stored virtual machine~(VM) snapshot to reset the environment. These function calls vary among tasks. And we summarize $5$ universal categories with their functionalities~(see Figure~\ref{fig:env_setup}), namely:
1)~\textit{File Transfer}: transfer files or project archives (either from local or cloud storage) into the VM;
2)~\textit{Application Launch}: open software on the desktop, \textit{e.g.,} Visual Studio Code and Chromium;
3)~\textit{Remote API Calls}: invoke tool-specific API calls for professional applications, especially those requiring authentic user accounts, to reset and configure cloud workspaces;
4)~\textit{Script Execution}: execute a shell script in VM to set up the initial state, \textit{e.g.,} run a Docker container to start a localhost webserver for {\tt Superset};
5)~\textit{Playwright Automation}: run web browser simulation with Playwright, \textit{e.g.,} sign into an account or click a specific button and redirect to the target web page. 

\subsection{Task-specific Evaluation}
\label{sec:evaluation}
\begin{figure}[htbp]
    \centering
    \includegraphics[width=0.98\textwidth]{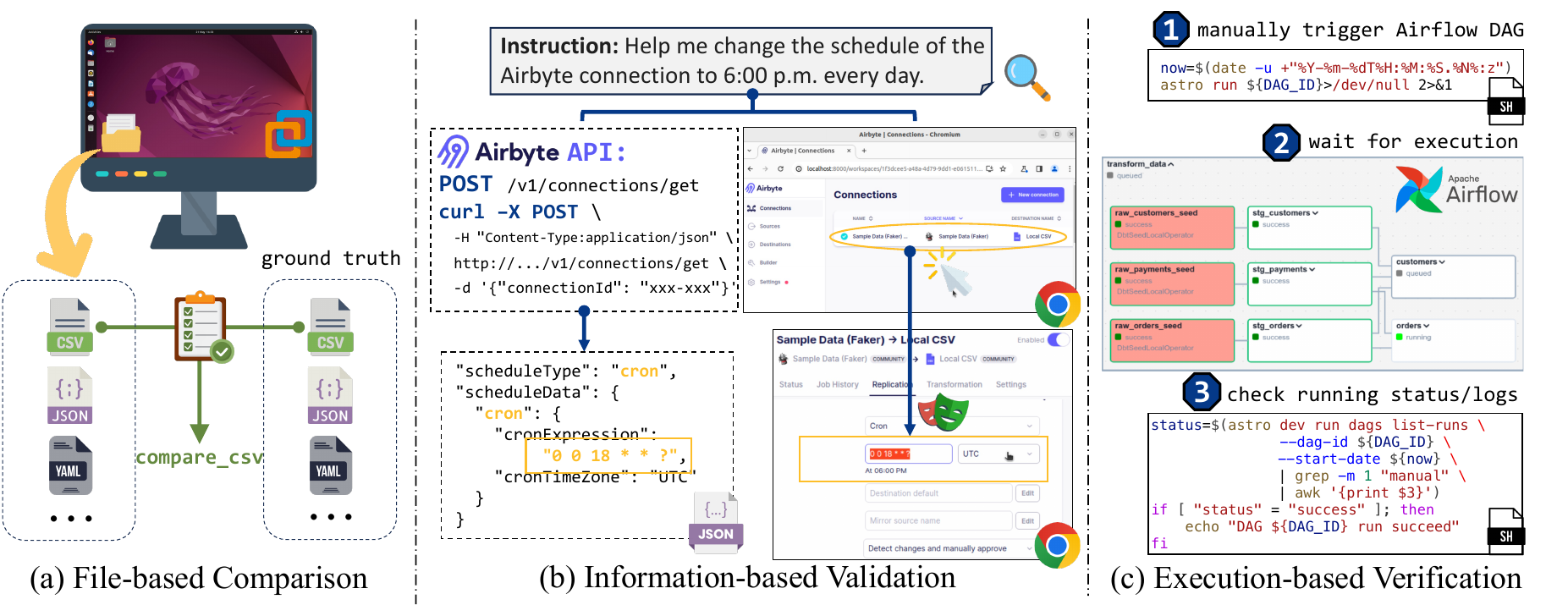}
    \caption{Three generic methods for task evaluation.}
    \label{fig:evaluation}
\end{figure}

After the interaction terminates, we only have access to the open-ended resulting state of the computer. Thus, to measure whether the goal of each task is accomplished, we write task-specific functions to retrieve the desired result from the open-ended resulting state and return the success flag~($0/1$). In total, \ours contains \initnum initial state configurations and \evalnum evaluation scripts, respectively. And we classify all evaluation methods into $3$ generic categories, also shown in Figure~\ref{fig:evaluation}:
\begin{enumerate}[leftmargin=*,label=\alph*)]
    \item \textit{File-based comparison}: this method finds and copies the target files from VM to the host, and resorts to file-type based metrics~(e.g., {\tt .json}, {\tt .csv}, etc.) to compare the specified aspect of the generated file with ground truth. Sometimes, the ground truth may be updated over time. In this case, we will fetch the latest labels from the Internet during evaluation.
    \item \textit{Information-based validation}: this scheme is usually utilized to extract and check desired information from the computer. For example, in Figure~\ref{fig:evaluation}(b), we want to confirm whether the time schedule of the data transportation is correctly configured in {\tt Airbyte}. We can invoke {\tt Airbyte} APIs to retrieve, or Chromium Playwright to locate the target value.
    \item \textit{Execution-based verification}: to verify whether an expected goal is achieved, we may also need to first execute a complicated Shell script in the final VM. For example, in Figure~\ref{fig:evaluation}(c), we manually trigger the target {\tt Airflow} DAG~\footnote{A DAG in {\tt Airflow} is defined as a collection of tasks to run, and DAG\_ID is used to uniquely identify it.} and check the eventual status through running logs.
\end{enumerate}


%% file: 3.benchmark.tex
\section{Benchmark Construction}
\label{sec:benchmark}

In this section, we introduce the general annotation pipeline, document warehouse construction, and dataset statistics for \ours. For concrete examples, refer to App.~\ref{app:examples}. 

\subsection{Annotation Pipeline}
\label{sec:annotation}

\begin{figure}[htbp]
    \centering
    \includegraphics[width=0.95\textwidth]{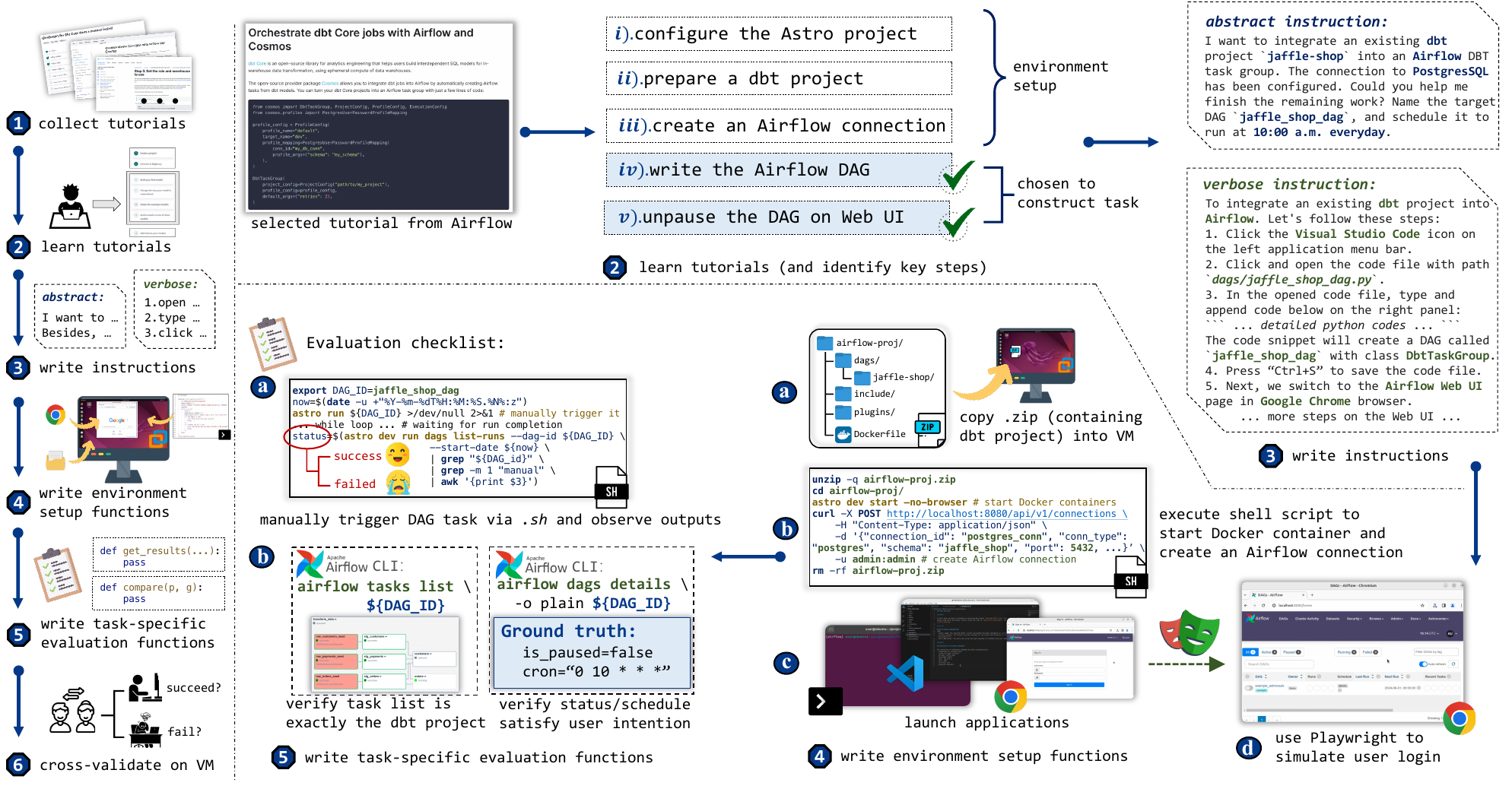}
    \caption{General annotation pipeline with one selected demonstration from the official {\tt Airflow} tutorial: \emph{Orchestrate dbt Core jobs with Airflow and Cosmos}.
    }
    \label{fig:procedure}
\end{figure}
To construct tasks in different categories, we find that official tutorials of enterprise applications serve as an excellent starting point. The $6$-step annotation pipeline is illustrated in Figure~\ref{fig:procedure}(a), and we elaborate it with a concrete and real example ``\emph{Orchestrate dbt Core jobs with Airflow and Cosmos}''~\footnote{The selected {\tt Airflow} tutorial URL: \url{https://www.astronomer.io/docs/learn/airflow-dbt}}:
\begin{enumerate}[leftmargin=*,label=\arabic*)]
    \item {\bf Collect tutorials:} firstly, we find tutorials from official websites for each professional tool in Figure~\ref{fig:task_categories}. In total, $10$ annotators collected \tutorialnum source URLs. Note that these tutorials may utilize other professional software, e.g., MySQL. All involved professional tools are listed in App.~\ref{app:softwares}.
    \item {\bf Learn tutorials:} the annotator selects one tutorial, learns and realizes it in the VM. After that, they can summarize key knowledge points from this tutorial. For example, in Figure~\ref{fig:procedure}(b), five key steps in integrating a {\tt dbt} project into an {\tt Airflow} task are extracted.
    \item {\bf Write instructions:} since the chosen tutorial is extremely complicated, the annotator can select a few key points to construct the task instruction. In Figure~\ref{fig:procedure}, we only select key steps \emph{iv)} and \emph{v)} to write two versions of instructions, \emph{abstract} and \emph{verbose}, indicating different levels of proficiency. Note that, to avoid potential data contamination and make the task more realistic, we ask the annotator to introduce at least two modifications to the raw tutorial. In this example, we a) replace the original ``{\tt my\_simple\_dbt\_project}'' into an open-source {\tt dbt} project called ``{\tt jaffle-shop}''~\footnote{URL of open-source {\tt dbt} project ``{\tt jaffle-shop}'':~\url{https://github.com/dbt-labs/jaffle-shop}}, and b) add one extra requirement on the time schedule (10:00 a.m. daily).
    \item \textbf{Write environment setup functions:} the next step is to write initialization functions using operations defined in~\cref{sec:env_setup}. In the example above, we need to: a) Upload an unfinished {\tt Airflow} project into the VM. 
    b) Execute a Shell script to launch the web server~(via Docker containers) for {\tt Airflow} under the project folder.
    c) Open all relevant applications on the desktop to simulate real user scenarios. 
    d) Use Playwright to auto-login to the default {\tt Airflow} account. 
    \item \textbf{Write task-specific evaluation functions:} In this step, annotators are required to programmatically obtain results from the open-ended states of VM and assess whether the task is completed using methods in~\cref{sec:evaluation}. In this example, the evaluator contains: a) manually run the target {\tt Airflow} DAG and verify the final status is ``success''; b) using {\tt Airflow} CLIs to retrieve details of the target {\tt Airflow} DAG, and compare {\tt dbt} sub-tasks, status and schedule with ground truth.
    \item \textbf{Cross-validate on VM:} to ensure correctness, we go through strict cross-validation. Each annotated task is sent to two other annotators to check: a) whether the chosen task reflects a real-world use case; b) whether verbose instruction accurately fulfills the task and its requirements in the abstract instruction; c) whether the environment can be reset to the same state in different trials; d) whether the evaluation is robust when we exactly follow the verbose instruction or modify some inconsequential steps; e) whether the evaluation score is $0$ if we deliberately make some mistakes~(red-teaming). The task is preserved only if it withstands all these tests.
\end{enumerate}
On average, the annotation of one task~(including cross-validation) costs roughly $4$ hours.

\subsection{Document Warehouse}
\label{sec:document}
Even senior data scientists query official documentation of professional applications when completing a complicated data engineering task. To compensate for the deficiencies of the data agents in utilizing enterprise professional software~(e.g., unaware of coding specifications or APIs), we build a document warehouse for \ours. Concretely, we recursively crawl the web pages from the root websites of the professional applications in Figure~\ref{fig:task_categories}. After pre-processing through heuristics~(refer to App.~\ref{app:documents}),
raw HTML web pages are convert into $3$ different formats for retrieval, namely a) pure text, b) markdown, and 3) simplified HTML. Eventually, we obtain \docnum documents in total.




\subsection{Dataset Statistics}

\begin{figure}[htbp]
\centering
\begin{minipage}{.5\textwidth}
    \centering
    \includegraphics[width=0.95\textwidth]{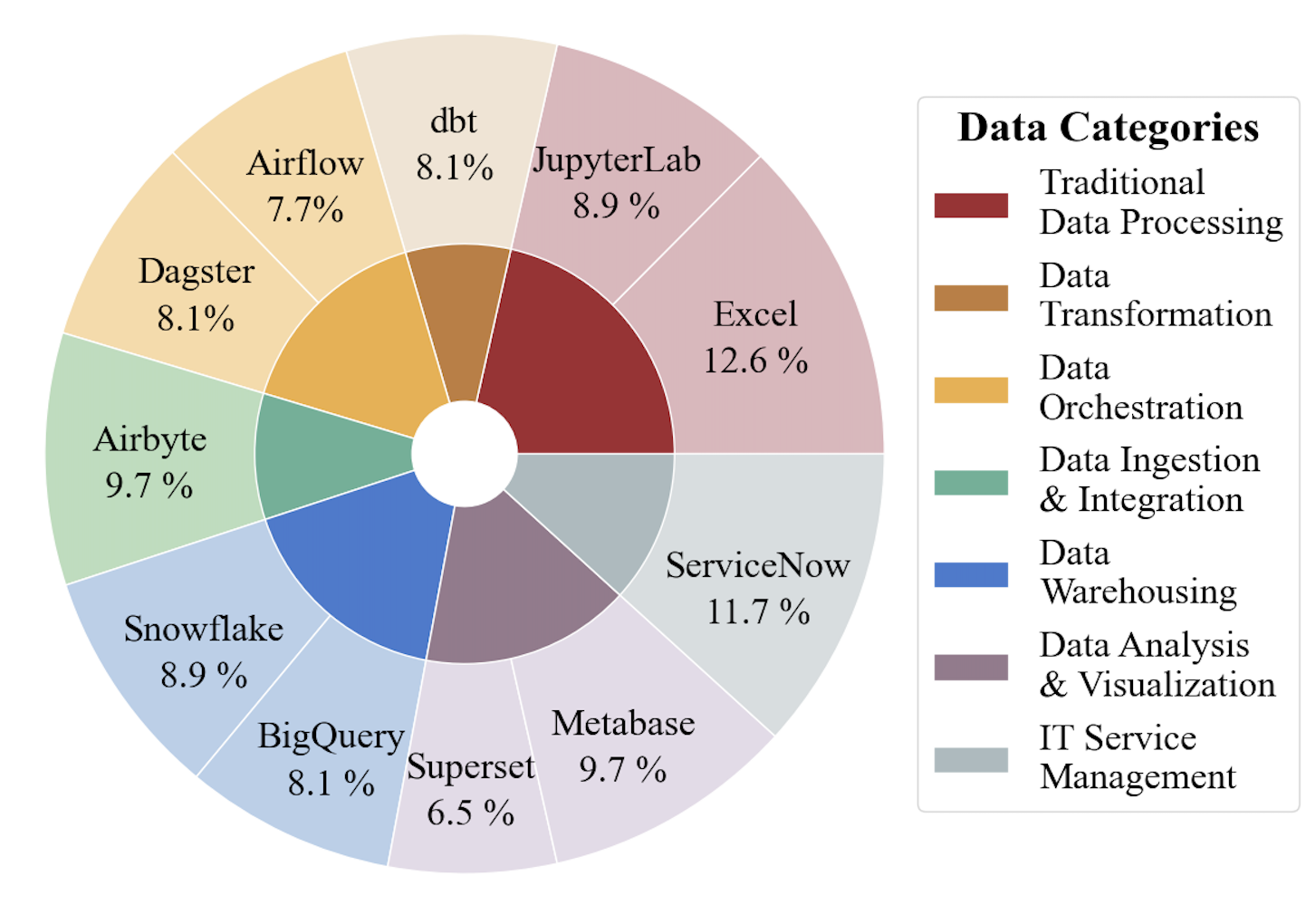}
    \caption{Task categories with professional tools.}
    \label{fig:task_categories}
\end{minipage}
\begin{minipage}{0.4\textwidth}
        \centering
            \captionof{table}{Statistics of \ours.}
		\scalebox{0.75}{
        \begin{tabular}{lc}
            \toprule
            \textbf{Statistics} & \textbf{Number} \\
            \midrule
            \textbf{Total Tasks} & \textbf{494 (100\%)} \\
            - Pure CLI & 28 (5.7\%) \\
            - Pure GUI  & 186 (37.7\%) \\
            - CLI + GUI & 280 (56.7\%) \\
            \midrule
             - w. Authentic User Account & 170 (34.4\%)\\
            - w/o. Authentic User Account & 324 (65.6\%) \\
            \midrule
            \textbf{Level (Action Steps)} &  \\
            - Easy~($\le 5$) & 98 (19.8\%) \\
            - Medium~($6\sim 15$) & 310 (62.8\%) \\
            - Hard~($>15$) & 86 (17.4\%) \\
            Avg. Action Steps & 4.0 / 9.6 / 22.0 \\
            \midrule
            Avg. Length of Abstract Instructions & 37.1\\
            Avg. Length of Verbose Instructions & 191.5\\
            Avg. Number of Used Apps Per Task & 2.5 \\
            \bottomrule
        \end{tabular}
        }
        \label{tab:key_statistics}
    \end{minipage}
\end{figure}


\begin{figure}[htbp]
\centering
\begin{minipage}{.32\textwidth}
    \centering
    \includegraphics[width=1.05\textwidth]{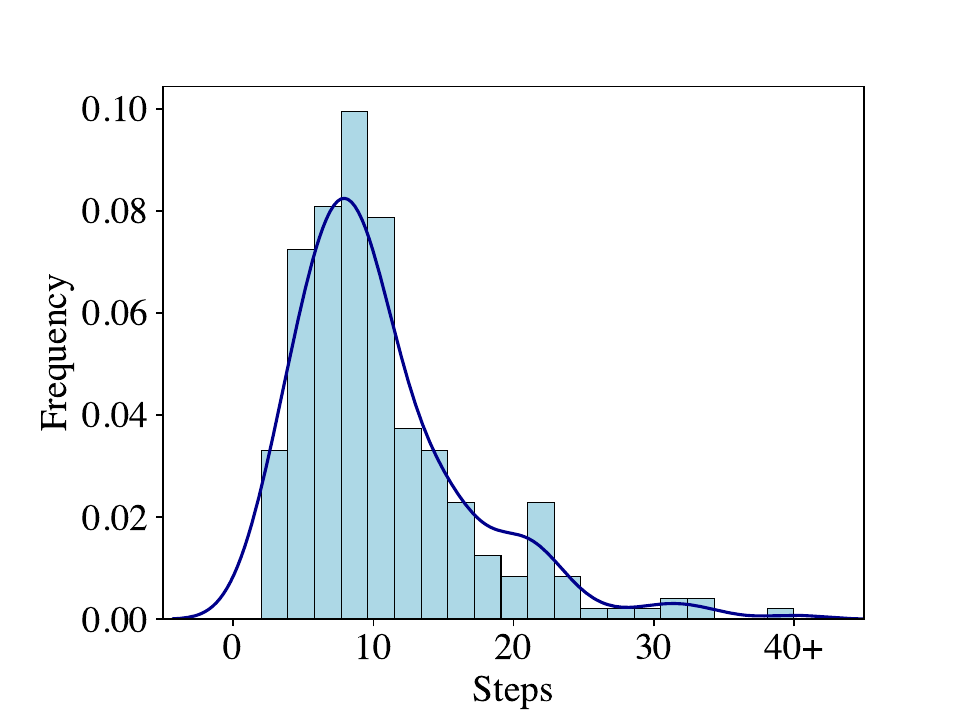}
\end{minipage}
\hspace{\fill}
\begin{minipage}{.32\textwidth}
    \centering
    \includegraphics[width=1.05\textwidth]{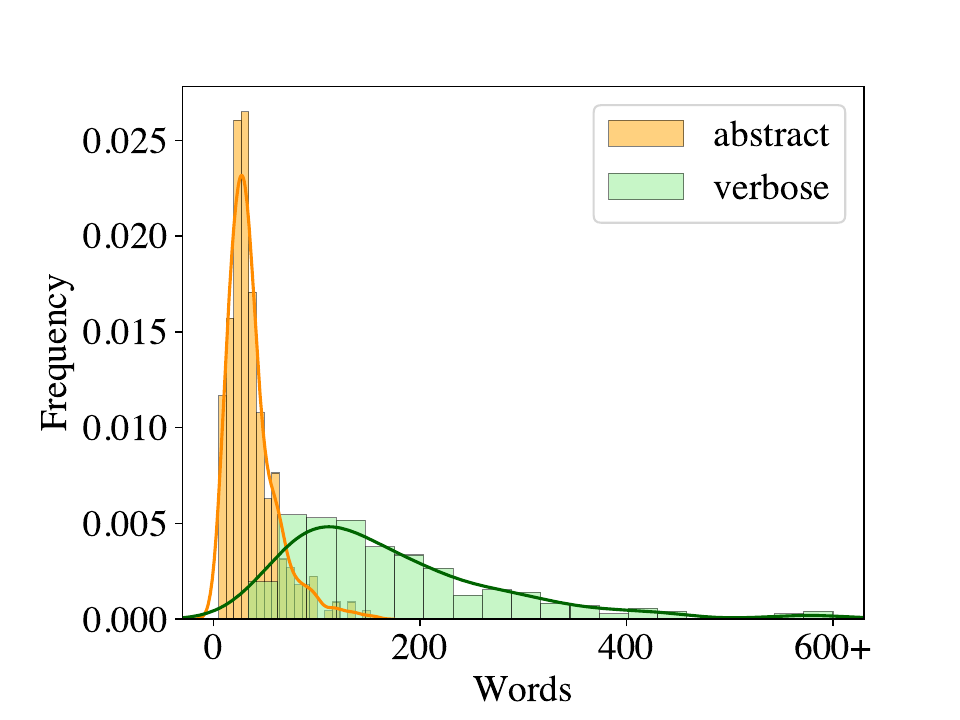}
\end{minipage}
\hspace{\fill}
\begin{minipage}{.32\textwidth}
    \centering
    \includegraphics[width=1.05\textwidth]{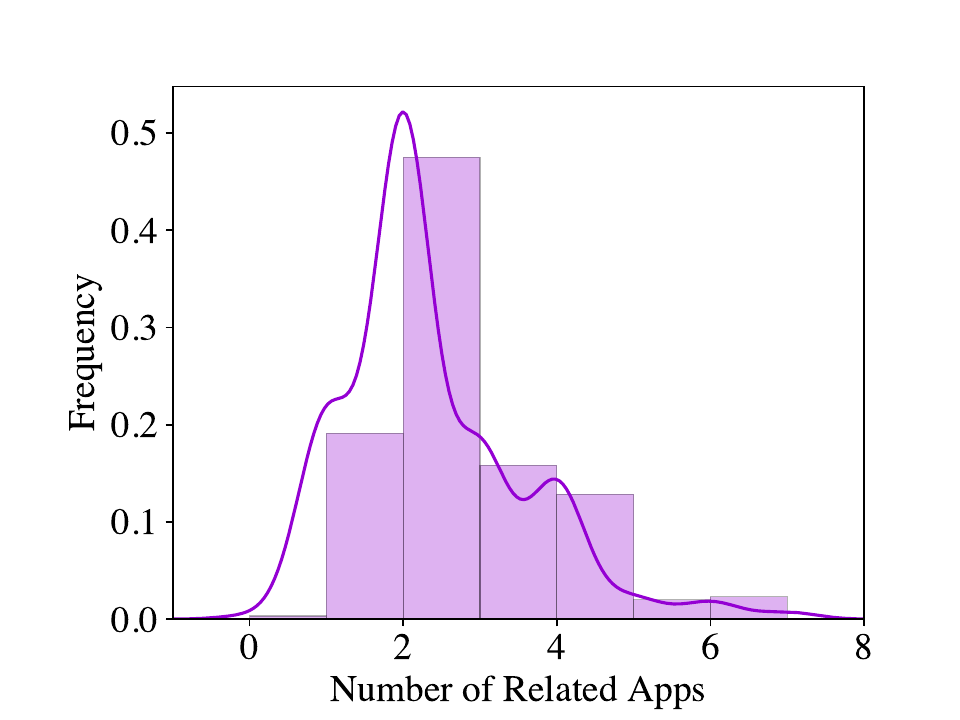}
\end{minipage}
\caption{Distribution of action steps, instruction length, and related applications per task.}
    \label{fig:stats}
\end{figure}

\paragraph{Tasks} We classify all $494$ tasks in \ours into $7$ categories and $11$ software sub-categories with main statistics in Figure \ref{fig:task_categories} and Table \ref{tab:key_statistics}. Specifically, most ($280$ tasks, $56.7\%$) involve CLI and GUI operations.
And $34\%$ examples request registering authentic software accounts.
Since each task is associated with a detailed, step-by-step tutorial~(verbose instruction), the entire task set can be categorized into three distinct levels based on the number of actions in these instructions.
The proportion of easy, medium, and hard tasks is approximately $1:2:1$.
According to the rightmost distribution depicted in Figure~\ref{fig:stats}, most tasks necessitate the coordinated utilization of multiple professional applications,
thereby establishing \ours as a particularly challenging benchmark.

\input{tables/benchmark_comparisons}

\paragraph{Comparison with existing benchmarks}
In Table \ref{tab:benchmark_comparisons}, we compare \ours with other agent benchmarks. \ours incorporates generic computer control commands into the field of data science and engineering and is distinguished by these salient features: 1) a real-time executable environment. Instead of providing static input-output pairs, \ours is equipped with a dynamic computer desktop such that agents can proactively explore it; 2) multiple enterprise software. We integrate $20$ professional applications into the benchmark, which include not only tools installed on local hosts but also cloud-based enterprise services; 3) intensive GUI operations. Unlike traditional coding or data science domains, experienced data scientists frequently manipulate the UIs of those professional software to simplify the data workflow~(\emph{e.g.}, enabling a specific function on the UI page or visualizing the graph view of data inputs).
In summary, \ours focuses on the use of professional enterprise software with visual interface in an interactive computer environment.

%% file: tables/benchmark_comparisons.tex
\begin{table}[htbp]
  \centering
  \caption{Comparison with existing agent benchmarks. Columns include the research field (Field), whether an executable environment is provided (Exec. Env.?), whether enterprise service is utilized (Ent. Serv.?), whether GUI actions are supported (GUI Support?) and some other statistics.
    }
  \renewcommand{\arraystretch}{1.2}
  \resizebox{0.98\textwidth}{!}{
    \begin{tabular}{lccccccccccc}
    \hline

    \hline
    \textbf{Benchmark} & \textbf{\thead{Field}}   & \textbf{\thead{Exec. \\ Env?}} & \textbf{\thead{Ent. \\ Serv.?}} & \textbf{\thead{GUI \\ Support?}} & \textbf{\thead{\# Apps/ \\ Sites}} & \textbf{\thead{\#  Exec.-based\\Eval. Func.}} & \textbf{\thead{ \# Tasks }} \\
    \hline \hline
    Spider \citep{yu2018spider}  & Text-to-SQL & \XSolidBrush & \XSolidBrush & \XSolidBrush & 1 & 0 &  1034 \\
    DS-1000 \citep{ds1000}  & Data Science  & \XSolidBrush & \XSolidBrush & \XSolidBrush & 1 & 0 &  1000 \\
    Arcade \citep{arcade} & Data Science & \XSolidBrush & \XSolidBrush & \XSolidBrush & 1 & 0 & 1082 \\
    MLAgentBench \citep{mlagentbench}  & Machine Learning  & \checkmark & \XSolidBrush & \XSolidBrush & 4 & 13 & 13 \\
    SWE-Bench \citep{swebench}         & Software Engineering   & \XSolidBrush & \XSolidBrush & \XSolidBrush & 12 & 1 & 2294 \\ 
    \hline
    Mind2Web \citep{deng2023mind2web}  & Web  & \XSolidBrush & \XSolidBrush & \checkmark & 137 & 0 & 2000 \\ 
    WEBLINX \citep{lu2024weblinx}      & Web  & \XSolidBrush & \XSolidBrush & \checkmark & 155 & 0 & 2337 \\
    WorkArena \citep{workarena}        & Web  & \checkmark & \checkmark & \checkmark & 1  & 7 & 29 \\
    AndroidWorld \citep{rawles2024androidworld} & Android  & \checkmark & \XSolidBrush & \checkmark & 20 & 6 & 116  \\     
    WebArena \citep{zhou2023webarena}  & Web & \checkmark & \XSolidBrush & \checkmark & 5 & 5 & 812 \\
    OSWorld \citep{osworld}            & Computer Control   & \checkmark  & \XSolidBrush & \checkmark & 9 & 134 & 369 \\ \hline
    \ours                              & \small{\makecell{Data Science \& Engineering \\ w/ Computer Control}} & \checkmark & \checkmark & \checkmark & 20 & 151 & 494 \\
    \hline

    \hline
    \end{tabular}%
    }
\label{tab:benchmark_comparisons}%
\end{table}%

%% file: 4.experiments.tex
\section{Experiments and Analysis}
In this section, we introduce the experiment settings, experimental results, and ablation study to assess the proficiency of current LLM or VLM based agents on \ours benchmark.
\subsection{Environment Settings}
\label{sec:agent_baseline}
\paragraph{Agent baselines} The baseline method includes $3$ schemes in zero-shot prompt learning: 1) Set-of-Mark~(SoM,~\citep{yang2023set}): following \textsc{OSWorld}~\citep{osworld} and VisualWebArena~\citep{Koh2024VisualWebArenaEM}, we adopt heuristic methods to retrieve coordinates of visible elements from {\tt a11ytree}~(a text-format observation type) and draw indexed bounding box for these elements on the screenshot. We further insert these indexes into the pruned {\tt a11ytree} to enhance the alignment between screenshot and {\tt a11ytree}. 2) Execution Feedback~(EF,~\citep{shinn2024reflexion}): we append execution feedback messages of those actions which failed to be grounded in the environment due to unexpected errors. The two techniques mentioned above are elaborated in App.~\ref{app:techniques}. 3) Retrieval-Augmented Generation~(RAG,~\citep{gao2023retrieval}): we leverage the task instruction as the query vector, {\tt bge-large-en-v1.5}~\citep{bge_embedding} as the embedding model, and LlamaIndex~\citep{Liu_LlamaIndex_2022} framework as the retrieval to generate document context for each task example.
Documents are pre-chunked into segments with maximum length $512$ and tokens overlapping size $20$. Top $4$ segments are selected as additional context in the task prompt~(detailed in App.~\ref{app:rag_prompt}).

\paragraph{LLMs and VLMs} We experiment with state-of-the-art LLMs and VLMs, including open-source representatives such as Mixtral-8x7B~\citep{jiang2024mixtral} and Llama-3-70B~\citep{meta2024llama3}, and closed-source ones including Qwen-Max~\citep{qwen}, Gemini-Pro-1.5~\citep{reid2024gemini}, Claude-3-Opus~\citep{claude3} and GPT~\citep{achiam2023gpt} families~(GPT-4o and GPT-4V~\footnote{We utilize the version {\tt gpt-4o-2024-05-13} for GPT-4o and {\tt gpt-4-1106-vision-preview} for GPT-4V.}). With respect to the two open-source LLMs and QWen-Max, we utilize pure text-format {\tt a11ytree} as the observation type on account of their incapability of image processing. For the remaining $4$ VLMs which support vision input, we use aligned text and image~(that is Set-of-Mark) as the observation type in main experiments. Unless otherwise specified, we set the temperature to $0.5$ and top\_p to $0.9$, the history trajectory window size to $3$, the maximum length of {\tt a11ytree} to $5000$ tokens, and the maximum output tokens to $1500$ in each turn. Heuristically, we require the agent to complete the tasks within both $15$ interaction turns and one hour, which suffices for most tasks~\footnote{Although some tasks require more than $15$ actions, we encourage the multimodal agent to predict multiple actions in one response in order to save the budget in the prompt design~(see App.~\ref{app:action_space_prompt}).}.

\subsection{Main Results}

\input{tables/main_table}
In Table~\ref{tab:main}, we compare performances of different LLMs and VLMs. All results above integrate techniques of both execution feedback~(EF) and retrieval-augmented generation~(RAG) in \cref{sec:agent_baseline}.
Accordingly, we can summarize that:
\begin{enumerate}[leftmargin=*,label=\arabic*)]
\item {\bf Existing data agents are far from satisfactory in completing real-world data science and engineering tasks.} Even state-of-the-art VLMs~(GPT-4o and GPT-4V) perform terribly on \ours, achieving at best \sota overall success rate. As for their strongest competitors, Gemini-Pro-1.5~\citep{reid2024gemini} and Claude-3-Opus~\citep{claude3}, they attain worse performances, even less than $10\%$ percents. There is still ample room for improvement in future work.
\item \textbf{Closed-source models are much more superior than open-source ones}. For those open-source LLMs, the success rate is exceedingly low, with some categories approaching zero. On one hand, it can be attributed to the fact that closed-source VLMs are pre-trained and fine-tuned on data of higher quality. On the other hand, closed-source VLMs support inputs with longer contexts and integrate both vision and text modalities~(further analyzed in \cref{sec:analysis}).
\item \textbf{Performances of data agents exhibit high variance, especially in categories ``\emph{data ingestion}'' and ``\emph{data visualization}''.} The majority of these two partitions are pure GUI tasks, which means agents mostly interact with the environment through time-dependent GUI operations. However, a minor error in one intermediate step can be amplified, resulting in the entire sequence of actions being wasted. Through error analysis on trajectories, we discover that once agents mispredict the coordinates of the correct button, they will open the wrong window and become trapped in the incorrect area, unable to return.
\item \textbf{Across $7$ data categories, the partitions ``\emph{data warehousing}'' and ``\emph{traditional data processing}'' are extremely challenging.} The reasons for this observation are two-fold: a) \emph{data warehousing} tasks mostly involve authentic user accounts~(\emph{e.g.}, {\tt BigQuery} and {\tt Snowflake}). Compared to other tasks which can be accomplished in a local host, these dynamic real-world scenarios incur extra burden on data agents, such as network connection delay and pop-up windows. Multimodal agents need to deal with these unexpected situations in real-time interaction with the computer. b) As for \emph{traditional data processing}, the bottleneck is that spreadsheets in Excel contain many cells, and it is particularly difficult for data agents to accurately locate the coordinates of cells. For example, applying the same math formula to the entire column requests multimodal agents to firstly pinpoint the right corner of a specific cell, wait for the mouse to become a cross, press and drag the mouse towards the target cell. This series of actions requires precise and fine-grained GUI controls which are difficult to implement.
\end{enumerate}

\subsection{Analysis}
\label{sec:analysis}
In this section, we delve into different factors which influence the eventual success rates, and analyze the underlying logics. The following analyses are based on our agent baseline with VLM GPT-4o unless otherwise specified. Firstly, we split the overall results into different subsets in Table~\ref{tab:task_subsets}.

\begin{enumerate}[leftmargin=*,label=\arabic*)]
\item {\bf Tasks with more inherent action steps are more difficult.}
Each task is associated with one verbose task instruction which gives a step-by-step guidance on how to complete it. We count the number of actions in the verbose instruction and split the entire task set into $3$ difficulty levels: $\le5$ steps~(Easy), $5\sim 15$ steps~(Medium), and $>15$ steps~(Hard). Not surprisingly, as the number of intrinsic action steps increases, the average performance decreases significantly. And for those extremely tough tasks, existing VLM-based data agents can hardly accomplish the goal.

\input{tables/ablation}
\item {\bf Tasks involving authentic user accounts are much more challenging.}
One salient feature of \ours is the integration of professional applications that require authentic user accounts. We also split the entire task set accordingly~(w/o or w/ account). Notably, data agents struggle to complete tasks involving authentic user accounts~($10.6\%$ success rate).
These tasks deal with real-world scenarios and incorporate cloud-hosted enterprise services.
Compared with Web servers which are launched locally in the VM~(\emph{e.g.}, from Docker containers), the cloud Web UIs 1) generally integrate more comprehensive functionalities or options in their menu panel, and 2) potentially suffer from emergency situation, such as extended network response delay due to bandwidth limitation or server overload. We conjecture these two causes collectively contribute to the inferior performances.

\item {\bf Incorporating GUI operations typically lead to improved performances.}
We split the task set by interfaces. If the task can be completed with pure CLIs~(e.g., code editor or bash terminal), we classify it as {\tt cli}. If the task only requires the agent to manipulate the GUI~(usually on the Web page), we classify it into {\tt gui}. For the remaining cases~({\tt cli+gui}), an agent must write code or scripts, and control the UI screen.
We observe that pure {\tt gui} tasks are much easier than {\tt cli} tasks. This conclusion can be explained by the following two reasons:
1) GUIs of professional applications are designed to simplify the original coding task. Clicking buttons or typing values on UIs can avoid handling the rigorous and complex coding specification.
2) Both observation types, namely the screenshot and \texttt{a11ytree}, are naturally proposed for GUI tasks. For pure {\tt cli} tasks, data agents must perform extra actions to locate and switch to the target panel before writing code.

\item {\bf Providing a step-by-step guideline in task instructions results in remarkable performance gains.}
The key difference between abstract and verbose instructions~(the third step in \cref{sec:annotation}) is whether a detailed step-by-step guidance is offered. With such stepwise oracle tutorials, data agents do not need to reason and plan, thus dramatically simplifying the original task. And the $4.8$ points improvement in Table~\ref{tab:task_subsets} consolidates this hypothesis. Nevertheless, the low success rate with verbose instructions~($16.2\%$) indicates that current VLMs still yield unsatisfactory results when purely grounding actions in real-world contexts. And significant potential remains for further enhancement.
\end{enumerate}

In Table~\ref{tab:abl}, we analyze the influence of different combinations of action space, observation types, and the $3$ techniques described \cref{sec:agent_baseline}. The findings include: 1) \textbf{Regarding action space, \textbf{\texttt{pyautogui}} code slightly outperforms self-customized JSON dict~($\mathbf{12.6\%}$ v.s. $\mathbf{10.5\%}$).} This can be attributed to the advantage that agents can also generate functional Python code like file traversal apart from the limited GUI control operations using the first action space. And it improves the efficiency of action grounding. 2) \textbf{As for observation types, single screenshot leads to very low performances~($\mathbf{4.2\%}$) on account of the agent's failure in pinpointing concrete elements.} When inserting {\tt a11ytree} into the observation which contains precise coordinates, the agent capability of locating target pixels is remarkably promoted. 3) \textbf{All $\mathbf{3}$ tricks we integrate into the agent baseline~(namely SoM, EF and RAG) will boost eventual performances.} It is interesting that if we do not adopt Set-of-Mark~(that is, enhancing the alignment between two modalities of observations), the result of {\tt screenshot+a11ytree} is even worse than that using pure {\tt a11ytree}. This emphasizes the significance of modal alignment when handling state observations.
\begin{wrapfigure}{r}{0.6\textwidth}
\centering
\begin{minipage}{0.6\textwidth}
    \centering
    \begin{subfigure}[b]{0.48\textwidth}
      \includegraphics[width=\textwidth]{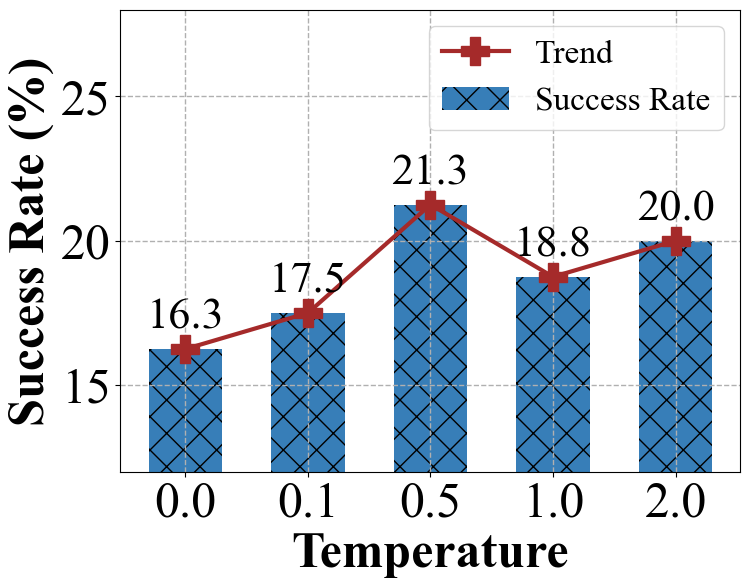}
    \end{subfigure}
    \hfill
    \begin{subfigure}[b]{0.48\textwidth}
      \includegraphics[width=\textwidth]{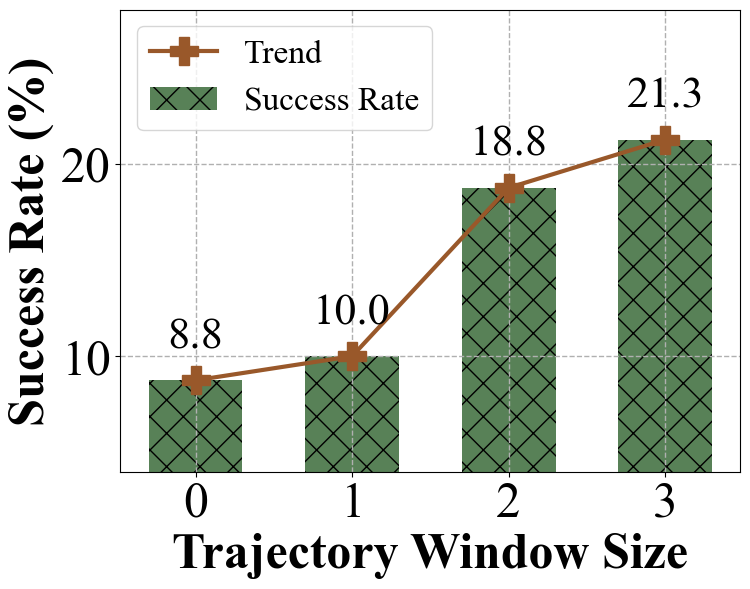}
    \end{subfigure}
    \caption{Ablation study on hyper-parameters.}
    \label{fig:ablation}
\end{minipage}
\end{wrapfigure}

\paragraph{A moderate temperature and longer history window size improve performances.} In Figure~\ref{fig:ablation}, we investigate the influences of two hyper-parameters on a task subset:
1) The top-ranked performance is achieved with sampling temperature $0.5$.
2) With the history window size enlarges, from $0$~(no history, only the current observation) to $3$, the performance increases stably. However, due to constraints on input length and considerations of cost-effectiveness, we are unable to extend the history trajectories any further. This also points out that the interaction efficiency is a serious issue and promising research direction.



%% file: tables/main_table.tex
\begin{table}[htbp]
  \centering
  \caption{Success rates of baseline agents on \ours grouped by $7$ task categories~(see Figure~\ref{fig:task_categories}), namely data warehousing~(\emph{ware.}), transformation~(\emph{trans.}), ingestion~(\emph{inges.}), visualization~(\emph{visual.}), orchestration~(\emph{orche.}), traditional data processing~(\emph{proc.}), and IT service management~(\emph{manag.}). For the first three LLMs, since they do not support visual information, we only utilize the text-based {\tt a11ytree} as the observation. For the remaining four VLMs, we adopt Set-of-Mark~(see \cref{sec:agent_baseline}).}
  \renewcommand{\arraystretch}{1.25}
  \resizebox{1.0\textwidth}{!}{
    \begin{tabular}{l|c|cccccccc}
    \hline

    \hline
    \multicolumn{1}{c|}{\multirow{2}{*}{\textbf{LLM / VLM}}} & \multirow{2}{*}{\textbf{Observation}} & \multicolumn{8}{c}{\textbf{Success Rate (\%)}} \\
\cline{3-10}       &       & \textit{ware.} & \textit{trans.} & \textit{inges.} & \textit{visual.} & \textit{orches.} & \textit{proc.} & \textit{serv.} & \textbf{Overall} \\
    \hline\hline
    Mixtral-8x7B &  \multirow{3}{*}{a11ytree} & $1.2$      & $0.0$      &  $0.0$     &   $0.0$    &  $2.6$     &  $0.9$     &  $0.0$   &  $0.8$ \\
    Llama-3-70B &    & $2.4$   &  $0.0$   &    $0.0$   &  $2.5$    &  $3.9$     &  $2.8$     &  $0.0$      & $2.0$   \\
    Qwen-Max &  & $1.2$ &   $0.0$   &  $0.0$     &  $0.0$     &  $2.6$     &  $0.0$     &   $0.0$    & $0.6$  \\
    \hline
Claude-3-Opus & \multirow{4}{*}{Set-of-Mark} &      $2.4$     &  $2.5$     &  $10.4$   &  $15.0$   & $11.5$  &  $3.8$     &  $12.1$ & $8.1$ \\
    Gemini-Pro-1.5 &     &  $3.6$  &  $2.5$     &  $14.6$     &  $15.0$     &  $10.3$     &   $2.8$    &  $\mathbf{19.0}$     &   $9.1$    \\
    GPT-4o &      &   $7.2$    &  $7.5$     &   $\mathbf{24.0}$    &   $14.1$    &   $\mathbf{19.8}$    &   $\mathbf{10.1}$    &    $13.8$   &  $13.8$ \\
    GPT-4V &   &  $\mathbf{10.8}$  &    $\mathbf{10.0}$   &    $12.0$   &   $\mathbf{25.0}$    &    $18.4$   &    $8.5$   &   $12.1$  &  $\mathbf{14.0}$ \\
    \hline

    \hline
    \end{tabular}%
    }
  \label{tab:main}%
\end{table}%

%% file: tables/ablation.tex
\begin{minipage}[t]{0.48\textwidth}
  \centering
  \captionof{table}{Success rate of GPT-4o with agent baseline SoM+EF+RAG across different partitions.}
    \label{tab:task_subsets}%
    \resizebox{0.8\textwidth}{!}{
    \begin{tabular}{l|cc}
    \hline
    
    \hline
    \textbf{Task Splits} & \textbf{Ratio~($\%$)} & \textbf{SR~($\%$)} \\
    \hline\hline
    Easy  & 19.8  & \textbf{38.8} \\
    Medium & 62.8  & 9.7  \\
    Hard  & 17.4  & 1.2  \\
    \hline
    w/o account & 66.0  & \textbf{15.6} \\
    w/ account & 34.0  & 10.6  \\
    \hline
    CLI   & 5.7   & 7.1  \\
    GUI   & 37.7  & \textbf{20.1} \\
    CIL+GUI & 56.7  & 10.6  \\
    \hline
    Abstract & 50 & 11.3 \\
    Verbose & 50 & \textbf{16.2} \\
    \hline

    \hline
    \end{tabular}%
    }
\end{minipage}
\hspace{\fill}
\begin{minipage}[t]{0.48\textwidth}
    \centering
    \captionof{table}{Ablation study on action space, observation types and $3$ tricks in \cref{sec:agent_baseline} on a task subset.}
    \label{tab:abl}%
    \resizebox{0.94\textwidth}{!}{
    \begin{tabular}{c|c|c}
    \hline

    \hline
    \textbf{\thead{Action \\ Space}} & \textbf{\thead{Observation \\ Types}} & \textbf{SR~(\%)} \\
    \hline\hline
    JSON dict & \multirow{2}{*}{screenshot} & 4.2 \\
    pyautogui &  & 4.2 \\
    \hline
    JSON dict & \multirow{2}{*}{a11ytree} & 10.5 \\
    pyautogui &  & \textbf{12.6} \\
    \hline
    \multirow{5}{*}{pyautogui} & screenshot+a11ytree & 11.4 \\
\cline{2-3}
& \multicolumn{1}{r|}{w/ Set-of-Mark} & 15.6 \\
          & \multicolumn{1}{r|}{w/ exec. feedback} & 13.6 \\
          & \multicolumn{1}{r|}{w/ retrieval aug.} & 14.4 \\
          & \multicolumn{1}{r|}{w/ all tricks} & \textbf{16.3} \\
    \hline

    \hline
    \end{tabular}%
    }
\end{minipage}

%% file: 5.related_work.tex
\section{Related Work}
\paragraph{Benchmarks for data science and engineering}
In the field of data science and engineering, several recent works propose novel benchmarks to evaluate the capabilities of LLM agents in manipulating Excel spreadsheets~\cite{sheetcopilot, sheetagent}, common data science libraries~(\emph{e.g.}, {\tt SQL} and {\tt pandas})~\citep{yu2018spider,ds1000,dabench,arcade}, machine learning~\citep{mlagentbench} or software engineering~\citep{sheetcopilot} projects. They are usually confined to a single stage within the entire data pipeline, predominantly data processing and analysis, thus overlooking other stages such as data warehousing and orchestration from a broader perspective.
Besides, like other coding-related datasets~\citep{intercode,su2024arks,spider}, they merely focus on the command line interface, neglecting the fact that enterprise software usually has rich graphical user interfaces~(GUIs). And data scientists often combine code programming with intensive GUI operations to fulfill a data workflow. 
To this end, \ours is proposed as the first-of-its-kind multimodal agent benchmark in the field of data science and engineering, which covers the entire data workflow and integrates visual interfaces.

\paragraph{Benchmarks for multimodal agents}
Existing works on GUI interaction mainly encompass web navigation~\citep{shi2017world,liu2018reinforcement,yao2022webshop,deng2023mind2web,Koh2024VisualWebArenaEM}, mobile device~\citep{zhang2023appagent,zhang2023mobile,rawles2023android, rawles2024androidworld,wang2024mobile}, and computer desktop~\citep{osworld,wu2024copilot,gao2023assistgui,Kapoor2024OmniACTAD}. One trend of recent advanced benchmarks is to provide an executable simulation environment. Multi-modal agents can explore and interact with this platform through keyboard, mouse, gesture and touch screen actions 
in a more realistic and complex scenario.
However, previous literature mostly focuses on daily life applications~(\emph{e.g.}, Web browser and calendar)~~\citep{xu2023tool,qin2023tool} or workflows of non-specialized business tasks~\citep{wornow2024multimodal}.
Few works~\citep{workarena,osworld,wornow2024multimodal} investigate the capability of multimodal agents to manipulate enterprise-level software. GUIs of professional applications often contain abundant domain-specific terminologies~(\emph{e.g.}, ``\emph{materialization}'' in {\tt Dagster}), which requires multimodal agents to understand the specialized knowledge. \ours incorporates $20$ professional tools into a real-time computer environment to test the proficiency of agents in data science and engineering. Furthermore, we supplement a large volume of documents for retrieval to compensate for deficiencies of agents in domain knowledge.

%% file: 6.conclusion.tex
\section{Conclusion}
In this work, we propose \ours, the first data science and engineering benchmark which integrates enterprise professional applications and supports intensive GUI operations besides code writing across the full data pipeline. It contains \tasknum tasks, involves \toolnum professional tools, and provides a real-time executable computer environment. The most advanced VLM~(\sotamodel) still performs poorly on \ours~(achieving \sota success rate), rendering it a very challenging benchmark.
Although current multimodal agents are still far from automating data workflows, \ours presents an easily accessible benchmark and lays the foundation for future research.

%% file: appendices/softwares.tex
\section{Checklist of All Professional Software in \ours}
\label{app:softwares}
In Table~\ref{tab:software}, we list all professional tools incorporated in the \ours benchmark, as well as their categories and descriptions.
\begin{table}[htbp]
  \centering
  \caption{Summary of all applications in \ours (label $\heartsuit$ means a real account is needed).}
  \label{tab:software}
  \renewcommand{\arraystretch}{1.5}
    \begin{tabular}{m{7.5em}|c|m{23em}}
    \hline

    \hline
            \textbf{Category} & \textbf{Software} & \textbf{Description} \\ \hline \hline
    \multirow{11}[0]{=}{Data Warehousing} & BigQuery$^\heartsuit$ & Fully-managed enterprise data warehouse service offered by Google Cloud Platform (GCP). It enables rapid processing and analysis of large datasets using SQL-like queries. \\ \cline{2-3}
        & Snowflake$^\heartsuit$ & Cloud-based data warehousing and analytics platform for large-scale data storage and processing, providing services to load, store, query, and analyze datasets at scale. \\ \cline{2-3}
        & MySQL & High-performance and scalable relational database management system (RDBMS) that is widely used and suited for fast data retrieval. \\ \cline{2-3}
        & PostgreSQL & RDBMS to store and manage large amounts of data with extensive additional features. \\ \cline{2-3}
        & DuckDB & Self-contained, serverless RDBMS with column-store architecture for fast analytical queries. \\ \cline{2-3}
        & SQLite & Another lightweight and serverless RDBMS that optimizes queries or transactions on individual rows. \\ \hline \hline
    Data Ingestion and Integration & Airbyte & Build connections to extract, transform, and load data from multiple sources to various destinations. \\ \hline \hline
    \multirow{3}[0]{=}{Data Transformation} & dbt & Framework to transform, test, and deploy data in warehouses. With dbt, users may define data models, transform raw data, and provide data quality checks. \\ \cline{2-3}
        & dbt-cloud$^\heartsuit$ & Cloud-based platform to model, transform and analyze data in a scalable and collaborative manner. \\ \hline \hline
    \multirow{4}[0]{=}{Data Analysis and Visualization} & Metabase & Business intelligence tool to create custom dashboards, reports, and analytics. It provides a simple and intuitive interface to ask questions and create visualizations. \\ \cline{2-3}
        & Superset & Enables users to make interactive dashboards. It can connect to various data sources and create visualizations to explore and analyze the data. \\ \hline \hline
    \multirow{3}[0]{=}{Data Orchestration} & Dagster & Platform for building, deploying, and scheduling data pipelines. It integrates data from various sources and manages data transformation jobs with dependencies. \\ \cline{2-3}
        & Airflow & Programmatically schedule and monitor workflows in the form of Directed Acyclic Graphs (DAGs). \\ \hline \hline
    \multirow{4}[0]{=}{Traditional Data Processing} &  JupyterLab & Interactive web environment for code and visualizations. It deals with notebooks containing live code and narrative text. \\  \cline{2-3}
        & Excel & Spreadsheet software that allows users to create and edit data in tables, charts, and formulas. We use the open-source LibreOffice Calc instead of Microsoft Excel in our environment. \\
        \hline \hline
    IT Service \newline Management & ServiceNow$^\heartsuit$ & Cloud-based IT service management platform that provides a suite of tools and features to streamline incident management, service catalog, asset management, and workflow automation.  \\ \hline \hline
    Daily Applications & \multicolumn{2}{l}{Docker, Chromium, Visual Studio Code, Bash Terminal} \\
    \hline

    \hline
    \end{tabular}%
\end{table}%

%% file: appendices/documents.tex
\section{Details of Document Warehouse}
\label{app:documents}

\subsection{Document Websites for Professional Tools}

Table~\ref{tab:docs} lists the official documentation websites corresponding to different software. We crawled only the English documentation from each official website and selected documents matching the version installed in our testing environment for download.
We used {\tt HTTrack}~\footnote{\url{https://www.httrack.com/}}, a free and easy-to-use offline browser utility, to download the HTML files to a local directory, building all directories recursively. We also retained the directory structure of each website, as we believe the path of each document can, to some extent, represent the document's purpose. For example, the HTML files under the path ``docs.getdbt.com/docs/deploy'' are about deploying dbt in production or staging environments. This crawling step resulted in a total of \fulldocnum HTML files.

\subsection{Filtering of HTML pages}

We further filtered the crawled HTML pages based on two criteria: irrelevant content to software usage and pages containing invalid content. For the former, we mainly judged whether the page contained content related to software usage based on its path and manually confirmed it. For example, pages under "author" on the website often relate to the website developer or development team rather than software usage. Additionally, we removed category-type pages that only contained navigation information. Furthermore, we filtered out pages based on the number of tokens obtained by whitespace tokenization. We mainly removed pages with token counts less than 100, as we found that these pages predominantly contained invalid information such as access failures, invalid links, or webpage redirections. For example, the official website of \texttt{Dagster} contained numerous links to unreleased versions of documents, all of which resulted in access failures. Therefore, after removal, the number of valid pages corresponding to \texttt{Dagster} decreased from 10,065 to 332. Finally, We obtained \docnum filtered HTML files~(see Table~\ref{tab:docs}).

\subsection{HTML Preprocessing}

HTML files contain a significant amount of content unrelated to the actual content of the webpage, such as ``\texttt{<script>}'', ``\texttt{<style>}'' tags, tag attributes, and developer comments. These parts may provide aesthetics to the page but are irrelevant to the document-level information. Additionally, they often occupy a large portion of the HTML file, making it excessively long for LLMs to input. To perform Retrieval-Augmented Generation (RAG) more efficiently and to help models better understand software documentation, we preprocessed these HTML files in three formats: plain text, HTML, and Markdown. These three formats of data and the original HTML files will be released to facilitate future research. The token statistics of all data formats are shown in Table~\ref{tab:docs_statistics}. We describe the preprocessing details below:

\noindent\textbf{Plain Text:} 
We used \texttt{BeautifulSoup4}~\footnote{\url{https://beautiful-soup-4.readthedocs.io/en/latest/}} to extract the textual elements from the HTML DOM~\footnote{The Document Object Model (DOM) is an interface that represents an HTML document as a tree structure, where each node is an object corresponding to a part of the document.} tree and connected these elements using ``\texttt{\textbackslash n}''. This method allows us to obtain the HTML content in the simplest manner, but losing the structural information of the HTML may affect the model's understanding of the webpage content.

\noindent\textbf{Simplified HTML:} 
We remove all sub-trees of the HTML DOM which do not contain textual elements. We also filter out all \textit{headers, footers, copyrights, forms, and iFrames.} We removed all HTML tag attributes since they mostly do not contain actual content or semantic information. Additionally, when a node in the HTML DOM tree has only one child node, we remove that node and directly connect its child node to its parent node. This effectively simplifies the structure and depth of the HTML. The simplified HTML preserves both the structure and content information of the original HTML with fewer tokens.

\noindent\textbf{Markdown:} We further used the \texttt{markdownify}~\footnote{\url{https://github.com/matthewwithanm/python-markdownify}} tool to convert the simplified HTML into Markdown format. Markdown format uses fewer tokens to represent structural information compared to HTML, striking a good balance between HTML and plain text formats. Moreover, since pure text includes a substantial number of newline characters used to concatenate text elements and some parts of the text content in markdown files are directly concatenated without these newlines, this results in a smaller average number of tokens in markdown files compared to the pure text format.

Concrete examples of these three formats are detailed in the task prompts~(see App.~\ref{app:rag_prompt}). In our pilot experiments~(see Table~\ref{tab:compare_rag}), we compare the performances using different formats of retrieved documents on a subset~($130$ task samples) of \ours. And pure text format outperforms the others.

\begin{table}[htbp]
  \centering
  \caption{Performances with different formats of retrieved documents on a subset of \ours.}
    \begin{tabular}{r|c}
    \hline

    \hline
    \multicolumn{1}{c|}{\textbf{RAG Format}} & \textbf{Success Rate (\%)} \\
    \hline\hline
    Pure Text & \textbf{16.92} \\
    Markdown Syntax & 15.38 \\
    Simplified HTML & 15.38 \\
    \hline

    \hline
    \end{tabular}%
  \label{tab:compare_rag}%
\end{table}%

\input{tables/docs}
\input{tables/doc_statistics}

\label{app:websites}

%% file: tables/docs.tex
\begin{table}[htbp]
\centering
\renewcommand{\arraystretch}{1.5}
\caption{Summary of software documentation. OrigPageNum: The number of all web pages we crawled from the documentation website. FilteredPageNum: The number of web pages obtained after filtering out irrelevant or invalid pages. }
\begin{tabular}{c|m{16.5em}|c|c}
\hline

\hline
\textbf{Software}          & \textbf{Documentation Website}                         & \textbf{OrigPageNum} & \textbf{FilteredPageNum} \\ \hline\hline
dbt/dbt-cloud & \url{https://docs.getdbt.com/}                       & 1192        & 1102            \\ \hline
Dagster        & \url{https://release-1-7-2.dagster.dagster-docs.io/} & 10065       & 332             \\ \hline
Airflow        & \url{https://docs.astronomer.io/}                    & 493         & 489             \\ \hline
\multirow{4}{*}{Airbyte}  & \multirow{1}{*}{\url{https://docs.airbyte.com/}}                                       & \multirow{4}{*}{958} & \multirow{4}{*}{859} 
\\ \cline{2-2}
                          & \url{https://airbyte.com/tutorials/} &                      &             
\\ \cline{2-2}
                          & \url{https://airbyte-public-api-docs.s3.us-east-2.amazonaws.com/rapidoc-api-docs.html} &                      &                      \\ \hline
Superset       & \url{https://superset.apache.org/docs/}              & 120         & 68              \\ \hline
\multirow{2}{*}{Metabase} & \url{https://www.metabase.com/docs/v0.49/ }                                            & \multirow{2}{*}{404} & \multirow{2}{*}{384} \\ \cline{2-2}
               & \url{https://www.metabase.com/learn/}                &             &                 \\ \hline
Snowflake     & \url{https://docs.snowflake.com/en/}                & 4436        & 4431            \\ \hline
Bigquery      & \url{https://cloud.google.com/bigquery/docs/}        & 1330        & 1328            \\ \hline
Jupyter        & \url{https://jupyterlab.readthedocs.io/en/4.1.x/}    & 2241        & 2238            \\ \hline\hline
\textbf{Total}        &   & \textbf{21239}        & \textbf{11231}            \\ \hline

\hline
\end{tabular}
\vspace{5pt}
\label{tab:docs}
\end{table}

%% file: tables/doc_statistics.tex
\begin{table}[htbp]
\centering
\caption{Average number of page tokens of different documentation formats. We used \texttt{TikToken}, a fast BPE tokenizer for use with OpenAI's models, to calculate the token count for gpt-3.5-turbo. }
\renewcommand{\arraystretch}{1.5}
\begin{tabular}{c|c|c|c|c}
\hline

\hline
\textbf{Software}          & \textbf{OrigHTML}                         & \textbf{PlainText} & \textbf{SimpHTML} & \textbf{Markdown} \\ \hline\hline
dbt/dbt-cloud &     17954           & 1669        & 2963   &  1510       \\ \hline
Dagster        & 131777 & 2615       & 4704 &     2290      \\ \hline
Airflow        &     35011          & 2007         & 3885    &  1829       \\ \hline
Airbyte  &       30124                       & 2448 & 4328 &2329
\\ \hline
Superset       &     10798         & 1398         & 2389 &    1415         \\ \hline
Metabase &          33523                 & 2288 & 4690 & 2333 \\ \hline
Snowflake     &     105155         & 1750        & 3342   &   1595      \\ \hline
Bigquery      &   103748    & 6245        & 11777    & 5718       \\ \hline
Jupyter        &  224153  & 11240        & 19917 &    6743        \\ \hline\hline
\textbf{Total}        &   \textbf{109119}  & \textbf{4273}        & \textbf{7789}   &     \textbf{3212}    \\ \hline

\hline
\end{tabular}
\label{tab:docs_statistics}
\end{table}

%% file: appendices/environment.tex
\section{Details of Executable Environment in \ours}
\label{app:environment}
In this section, we briefly introduce \textsc{OSWorld}~\citep{osworld} and how we adapt it to meet our requirements.

\begin{figure}[htbp]
    \centering
    \includegraphics[width=0.9\textwidth]{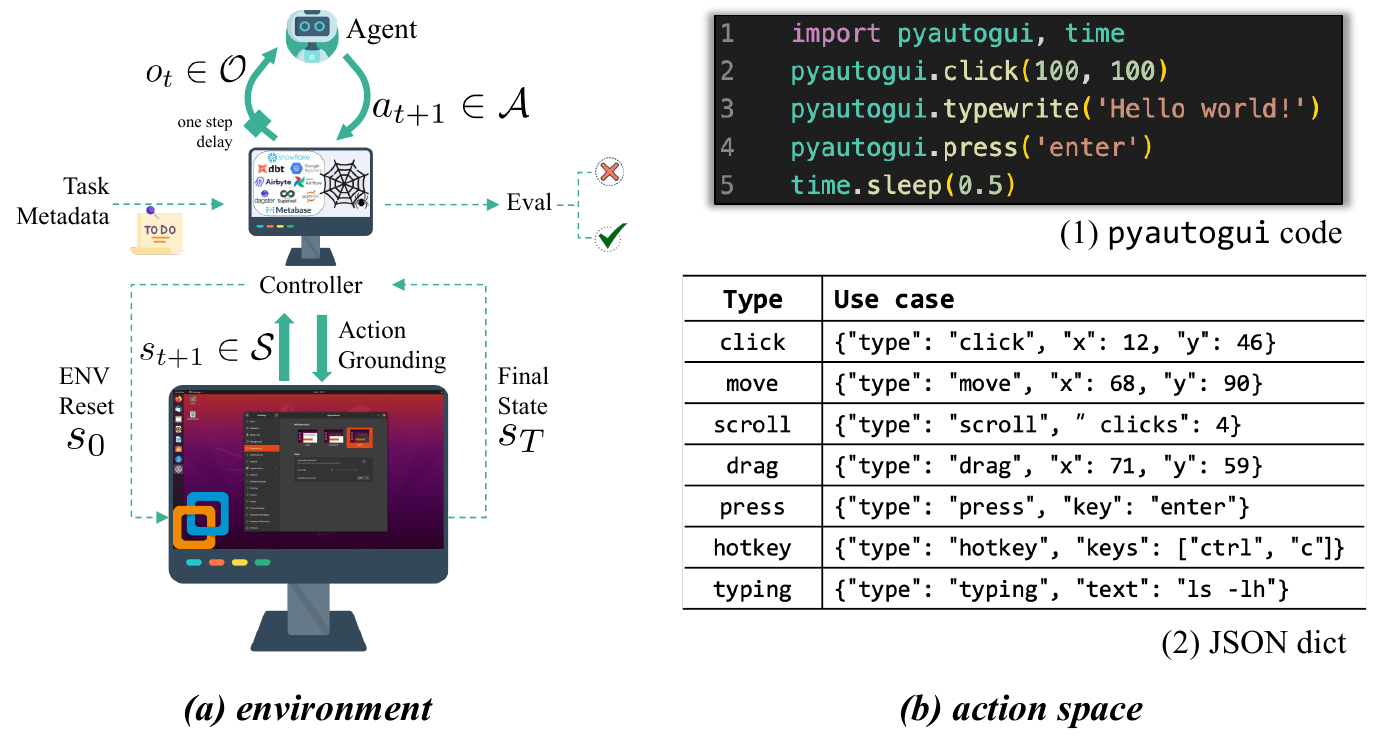}
    \caption{Overview of the executable environment of \ours and two types of action space. }
    \label{fig:app_environment}
\end{figure}

\subsection{Overview}
\ours formalizes the interaction with a Ubuntu desktop as a partially observable Markov decision process~(POMDP) $(\mathcal{S},\mathcal{O},\mathcal{A},\mathcal{T},\mathcal{R})$ with state space $\mathcal{S}$, observation space $\mathcal{O}$, action space $\mathcal{A}$, state transition function $T:\mathcal{S}\times\mathcal{A}\rightarrow\mathcal{S}$ and reward function $\mathcal{R}:\mathcal{S}\times\mathcal{A}\rightarrow\mathbb{R}$. Given the current observation $o_t\in\mathcal{O}$ from the desktop, the agent needs to predict action $a_{t+1}\in\mathcal{A}$ for the next step. An admissible action incurs a change in the latent state space $s_{t+1}\in\mathcal{S}$, and the environment feedback $o_{t+1}$. The interaction loop repeats until a special ``{\tt DONE}'' or ``{\tt FAIL}'' action is issued, wherein the task episode ends and a reward $r = \mathcal{R}(s_T)\in \{0, 1\}$ is computed, with $1$ indicating task success.

The executable computer environment~(a Ubuntu operating system) is built upon virtual machines~(VMs). By using the ``\emph{snapshot}'' functionality of VM, the localhost environment state can be completely recovered to a stored history state. This snapshot with task-specific setup functions~(see \cref{sec:env_setup}) serve as the initial state $s_0\in\mathcal{S}$ for different tasks. And a core \emph{controller} is responsible for grounding action $a_t$~(see App.~\ref{sec:action}) into the VM desktop and obtaining observations $o_t$~(see App.~\ref{sec:observation}) from the resulting state of VM. After the agent issues a special action ``{\tt DONE}'' or ``{\tt FAIL}'', the controller will invoke the customized evaluation function for the current task~(see \cref{sec:evaluation}) and report the metric score. The entire procedure is shown in Figure~\ref{fig:app_environment}(a).

\subsection{Action Space}
\label{sec:action}
For generic actions that support both CLI and GUI, we introduce two different action spaces:

\paragraph{pyautogui code} This action space accepts arbitrary executable python code. Particularly, code snippets that using python library ``{\tt pyautogui}'' to control the mouse and keyboard are strongly recommended. Generally, mouse-based actions~(\emph{e.g.}, click and scroll) directly manipulate the GUI screen, while keyboard-based actions~(\emph{e.g.}, typewrite and hotkey) interact with the CLI such as the bash terminal and code editor~(\emph{e.g.}, Visual Studio Code). 

\paragraph{JSON dict} Inspired by the ``{\tt pyautogui}'' library, we summarize $7$ actions to simplify the action space. This small set can cover all CLI and GUI actions needed on the desktop. For each action and its parameters, we further encapsulate it into a JSON dict to restrict the output format. The API specification and use cases are formally described in prompt messages~(see App.~\ref{app:action_space_prompt}). And the checklist of all $7$ actions is presented in Figure~\ref{fig:app_environment}(b).

\subsection{Observation Space}
\label{sec:observation}
\begin{figure}[htbp]
    \centering
    \includegraphics[width=0.95\textwidth]{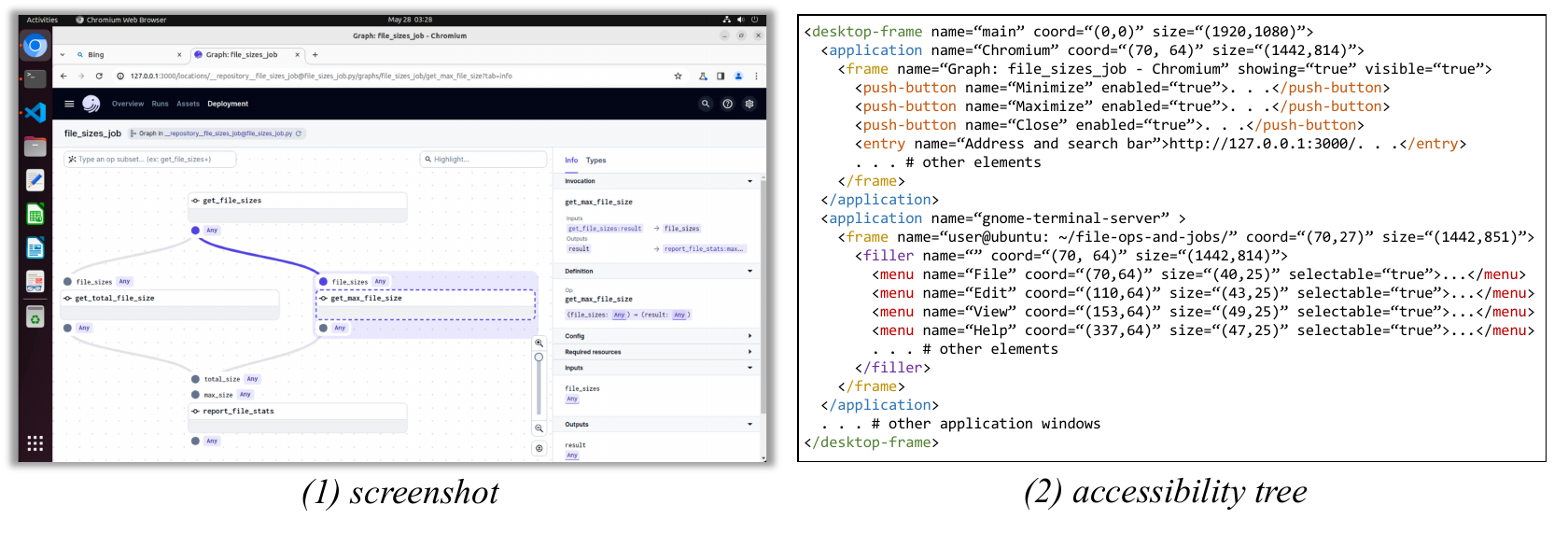}
    \caption{Two observation types: screenshot and accessibility tree~({\tt a11ytree}).}
    \label{fig:app_observation}
\end{figure}
With respect to observations, there are two widely used alternatives~(see Figure~\ref{fig:app_observation}): 1) image-style screenshot of the entire desktop, and 2) text-format accessibility tree~({\tt a11ytree}). The accessibility tree, obtained from the Assistive Technology Service Provider Interface (ATSPI) library~\footnote{\url{https://docs.gtk.org/atspi2/}}, is a text-format abstraction of the entire computer desktop which describes the name, type, status~(\emph{e.g.}, a menu bar is ``\emph{selectable}''), position~(\emph{e.g.}, in Figure~\ref{fig:app_observation} (2), the attributes ``\emph{coord}'' and ``\emph{size}'' together define the rectangle position), and text content embedded in each element~(e.g., windows, panels, buttons, and input boxes). We extract {\tt a11ytree} using python library {\tt pyatspi} and convert it into the XML format. It functions similar to DOM~(Document Object Model) tree for websites.

\subsubsection{Two tricks: Set-of-Mark and Execution Feedback}
\label{app:techniques}
\begin{figure}[htbp]
    \centering
    \begin{minipage}{0.48\textwidth}
        \includegraphics[width=\textwidth]{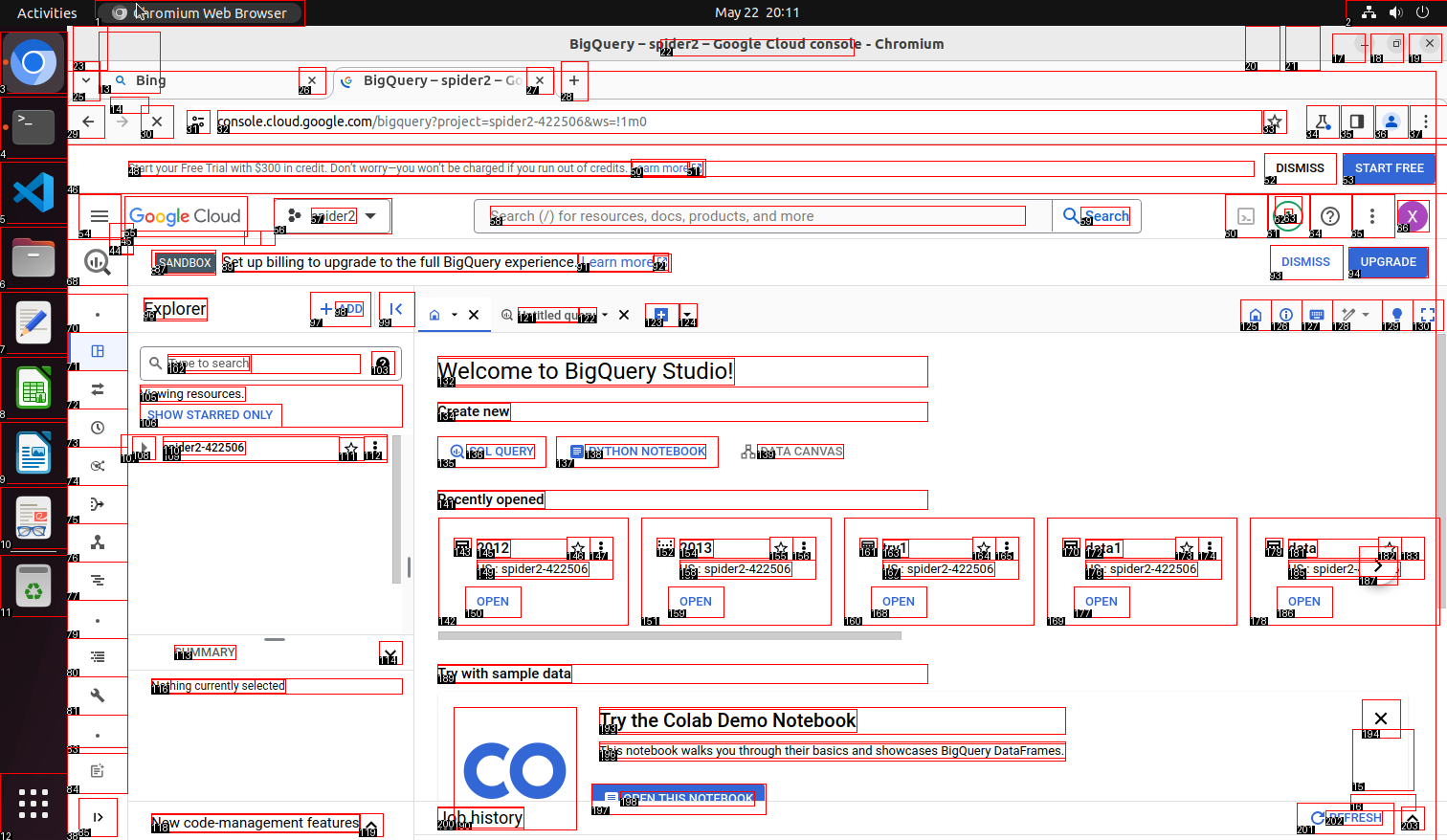}
        \caption{Screenshot with bounding boxes.}
        \label{fig:som_screenshot}
    \end{minipage}
    \hfill
    \begin{minipage}{0.48\textwidth}
        \includegraphics[width=\textwidth]{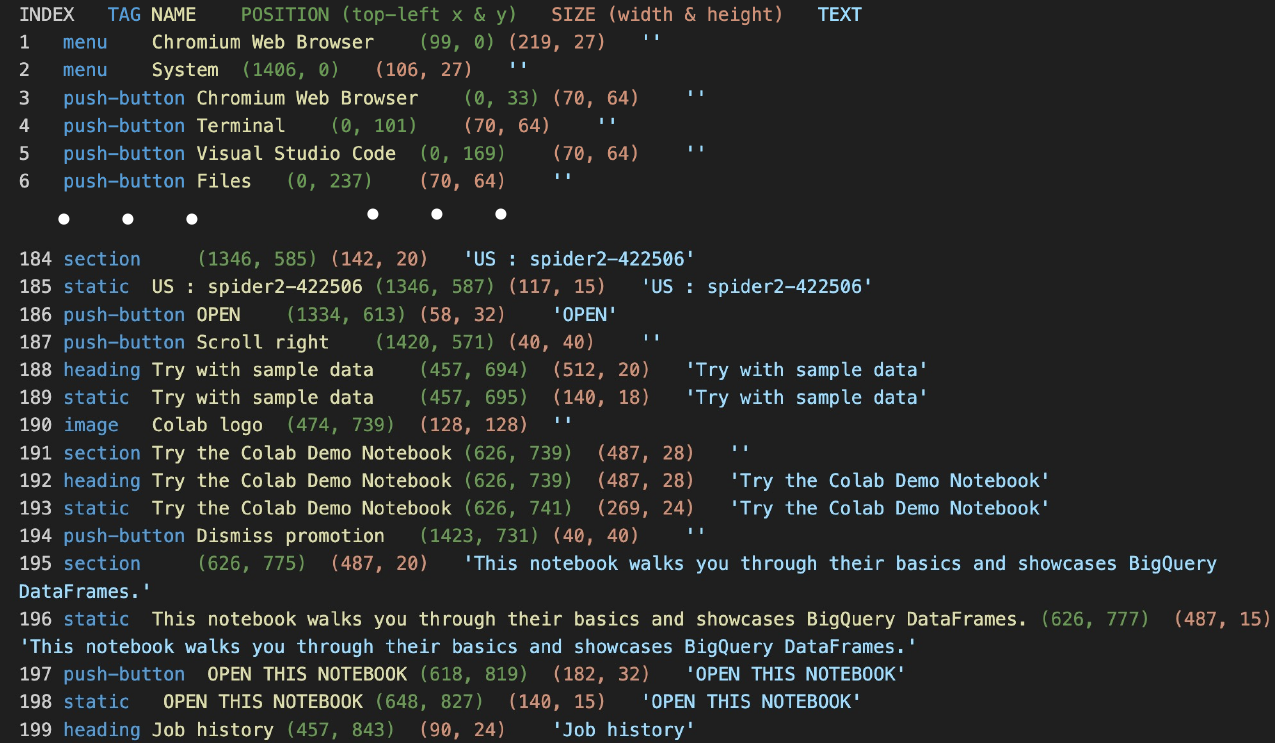}
        \caption{Converted table of {\tt a11ytree}.}
        \label{fig:som_a11ytree}
    \end{minipage}    
    \caption{Illustration of the aligned observation type set-of-mark~(SoM).}
    \label{fig:som}
\end{figure}
\paragraph{Set-of-Mark~(SoM)} The original text-style accessibility tree~({\tt a11ytree}) and image-style screenshot do not align with each other. To compensate for this deficiency, we follow \textsc{OSWorld}~\citep{osworld} and WebArena~\citep{zhou2023webarena} to draw bounding boxes for elements of interest in the screenshot and label these elements with numeric indexes. The accurate coordinates of these bounding boxes are extracted from the {\tt a11ytree}. Furthermore, we re-organize the {\tt a11ytree} into a table~(each leaf node in {\tt a11ytree} is converted into one row) and insert another attribute/column ``{\tt index}'' for each node in the tree. The value of attribute ``{\tt index}'' is exactly the numeric label of the corresponding element in the screenshot. The aligned screenshot and {\tt a11ytree}~({\it a.k.a.}, set-of-mark, SoM~\citep{yang2023set}) are illustrated in Figure~\ref{fig:som}.

\begin{tcolorbox}[label=ef, title={Examples of Execution Feedback Messages}]
\begin{Verbatim}[commandchars=!\{\}]
Here are failed actions with their error messages in your last response:
!textcolor{purple}{# Action 1}
import pyautogui
index_34 = (23, 43)
pyautogui.click(index_343)
!textcolor{blue}{# Execution error:}
Traceback (most recent call last):
NameError: name 'index_343' is not defined

!textcolor{purple}{# Action 2}
import pyautogui
import time
pyautogui.write('USE DATABASE IMDB\n\\n')
!textcolor{blue}{# Execution error:}
File "<string>" line 3
pyautogui.write('USE DATABASE IMDB
                ^
SyntaxError: unterminated string literal
\end{Verbatim}
\end{tcolorbox}

\paragraph{Execution Feedback} We also incorporate another type of information as the observation, namely the {\it execution feedback} of actions~(see messages above). We notice that, some predicted actions may be parsed erroneously or fail to be executed. In this case, the two observation types mentioned before are not changed at all. And the agent repeatedly urges to conduct the same incorrect action. To inform the agent of execution errors, we include this execution feedback as the third observation type.

%% file: appendices/task_example.tex
\section{Format of Task Examples}
\begin{figure}[htbp]
    \centering
    \includegraphics[width=0.98\textwidth]{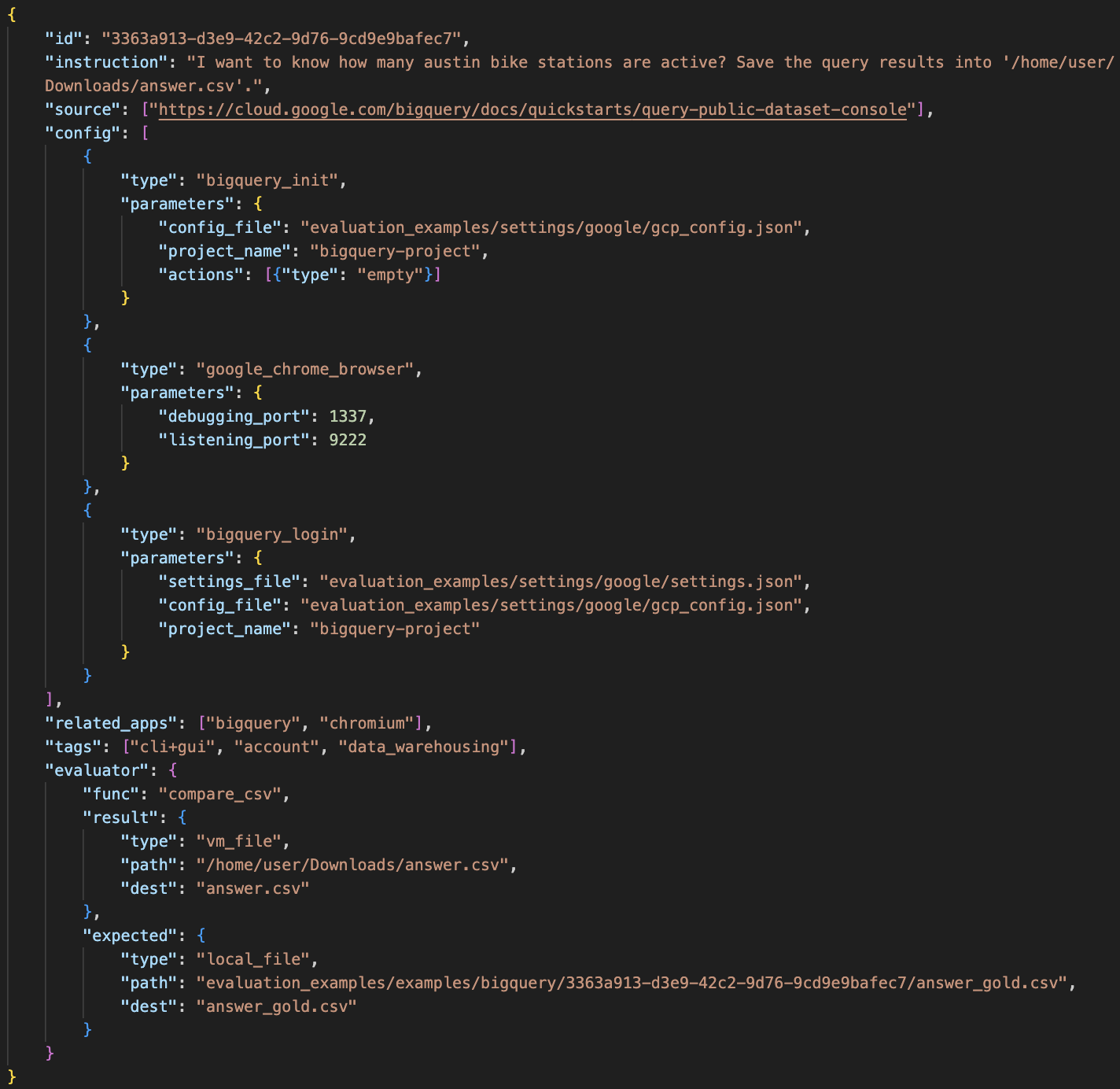}
    \caption{The format of a simple task example~({\tt .json} configuration file).}
    \label{fig:task_example}
\end{figure}

In this section, we briefly introduce the format of task examples. Following \textsc{OSWorld}~\cite{osworld}, each task instance is represented as a JSON dictionary which contains the following fields: (see Figure~\ref{fig:task_example})
\begin{itemize}
    \item {\tt id}: globally unique id of the current task example.
    \item {\tt instruction}: the task instruction which indicates the task goal.
    \item {\tt source}: a list of referenced tutorial links to construct the current task.
    \item {\tt config}: a list of dictionaries which define the operations to initialize and reset the computer desktop. Each dictionary contains the function name~(the ``{\tt type}'' key) and its parameters~(the ``{\tt parameters}'' key) indicating one environment setup function. For example, in Figure~\ref{fig:task_example}, we define $3$ environment reset functions, namely 1) ``{\tt bigquery\_init}'' to clear the cloud workspace of Google project ``{\tt bigquery-project}'', 2) ``{\tt google\_chrome\_browser}'' to launch the Google Chrome application, and 3) ``{\tt bigquery\_login}'' to simulate the Google account login operation with playwright.
    \item {\tt related\_apps}: a list of application names which should be used in the current task.
    \item {\tt tags}: a list of tags denoting different categories.
    \item {\tt evaluator}: a dictionary containing $3$ fields: {\tt func}, {\tt result}, {\tt expected}. It defines how to evaluate the final results once task completion. Concretely, the ``{\tt func}'' field defines the name of our customized function~(or metric) which is used to compare the predicted result and the expected golden result. The ``{\tt result}'' field defines how to extract the predicted result from the final environment states. And the ``{\tt expected}'' field defines how to obtain the golden result. For example, in Figure~\ref{fig:task_example}, we utilize the function ``{\tt compare\_csv}'' to compare the predicted file ``{\tt /home/user/Downloads/answer.csv}'' in the virtual machine and the golden file ``{\tt answer\_gold.csv}'' in local host.
\end{itemize}

%% file: appendices/examples.tex
\clearpage
\section{Task Examples}
\label{app:examples}
In this part, we present diverse examples in \ours.

\begin{longtable}{m{1.5cm}m{6cm}m{6cm}}
\caption{Real task examples from \ours.}
\label{tab:examples} \\\hline

\hline
\textbf{Related App(s)} & \textbf{Instruction} & \textbf{Screenshot After Initialization} \\
\hline
\endfirsthead

\multicolumn{3}{c}%
{{\tablename\ \thetable{} -- continued from previous page}} \\

\hline
\textbf{Related App~(s)} & \textbf{Instruction} & \textbf{Screenshot After Initialization} \\
\hline
\endhead

\multicolumn{3}{r}{{\textit{Continued on next page}}} \\ 
\endfoot

\endlastfoot

Dagster
dbt
Chromium
VS Code & \textit{I have a dbt project "jaffle\_shop". Please integrate this project into dagster and add a dagster asset "customers" according to the schema provided by the file "\textasciitilde/dbt-dagster-project/jaffle\_shop/customers\_schema.yml". Materialize the asset in the opened dagster UI.} & \includegraphics[width=6cm, height=3.38cm]{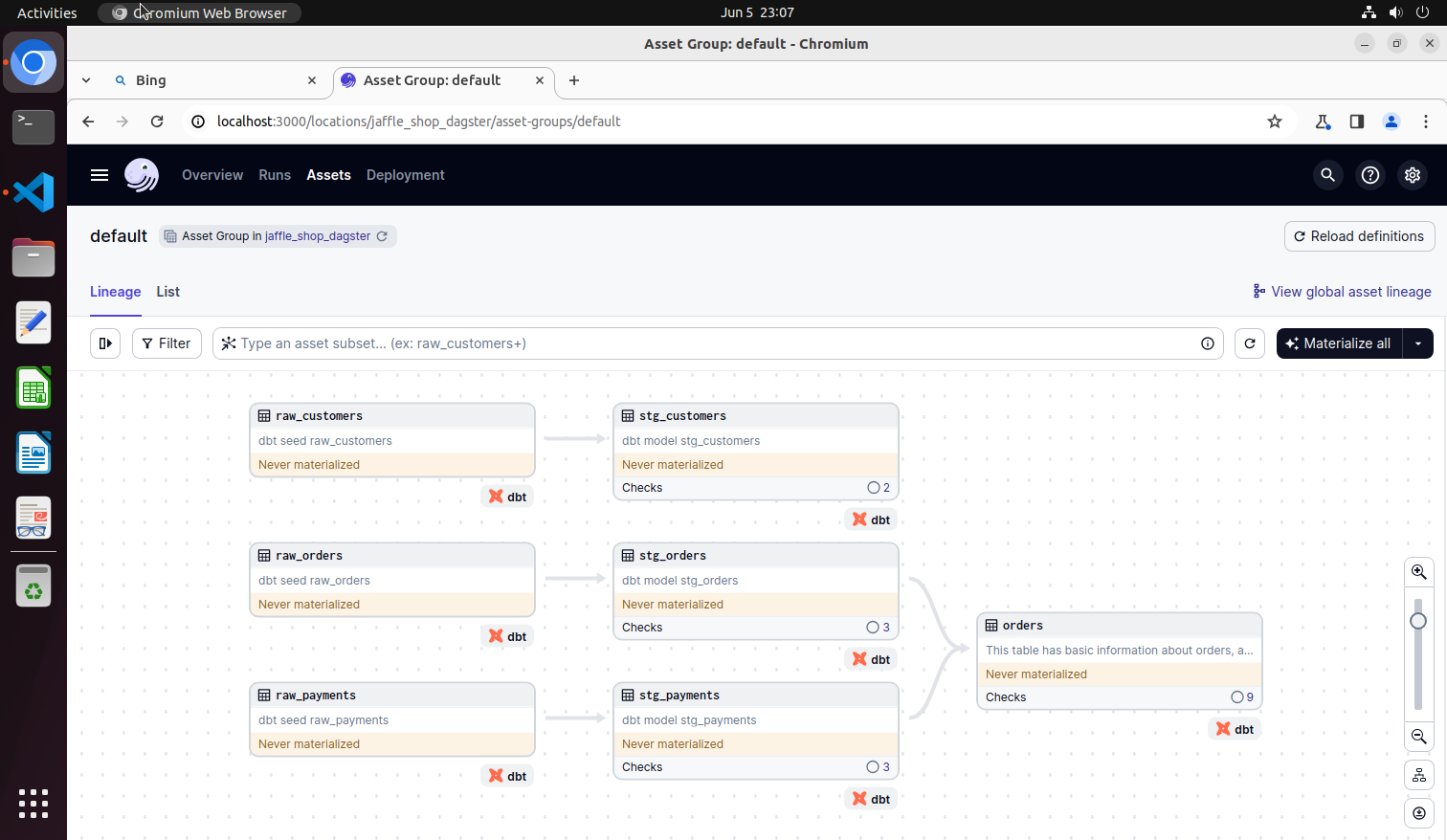}
\\\hline
BigQuery
Chromium & \textit{I have just uploaded data about Ameraican babies into table `names\_2014`. I am curious about the top five names for US babies that were assigned male at birth in that year. Please save the `name` and `count` into another table `top5\_male\_2014` in the same dataset for me.} & \includegraphics[width=6cm, height=3.38cm]{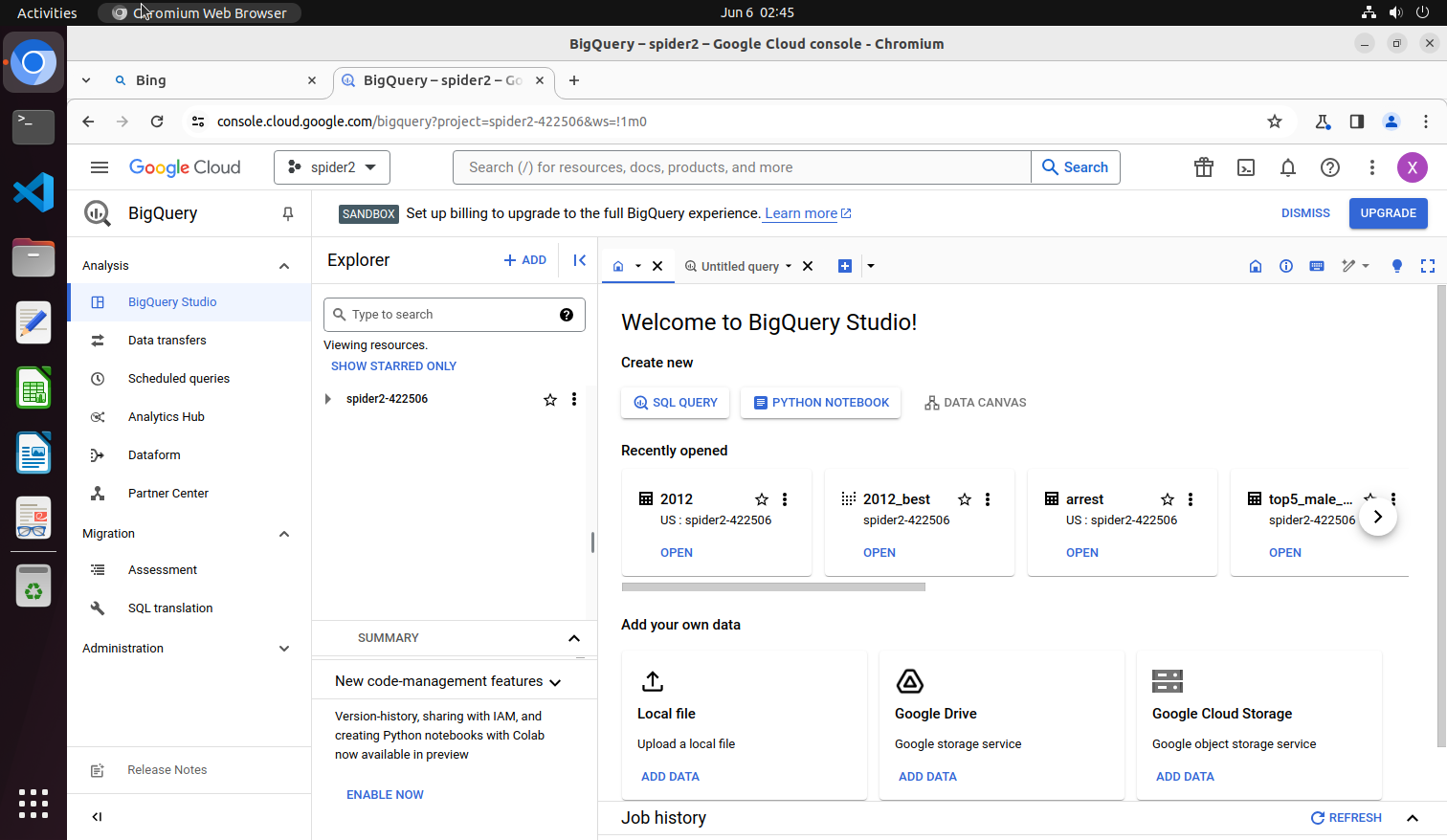}
\\\hline
Dagster
Airflow
MySQL
Chromium
VS Code\newline
Terminal & \textit{I have defined an Airflow DAG. Please help me migrate it to Dagster based on the requirements in "README.md". Remember to launch the Dagster webserver from "dagster\_migration.py" and start the DAG schedule. Test the schedule on Dagster UI Launchpad and make sure the job can succeed.} & \includegraphics[width=6cm, height=3.38cm]{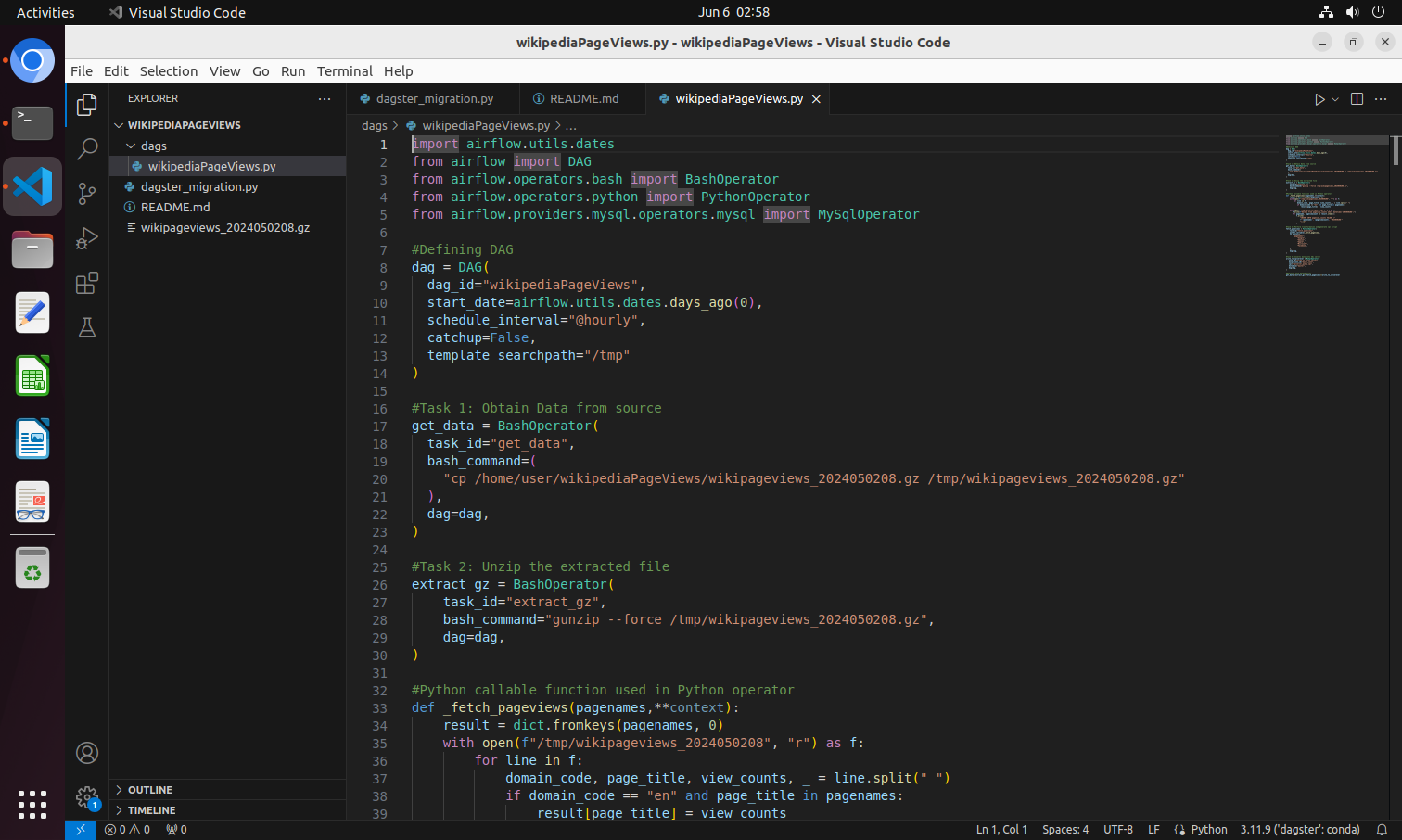} \\\hline
Metabase
Chromium & \textit{I want to have a stack bar chart out of Sample Database in metabase. Could you help me visualize the data of Products table and summarize the data of Sum of price by Product Category and Created At - Quarter. Then stack the visualized chart. Please help me download the visualization as a PNG file, and rename it to "stack\_chart.png".} & \includegraphics[width=6cm, height=3.38cm]{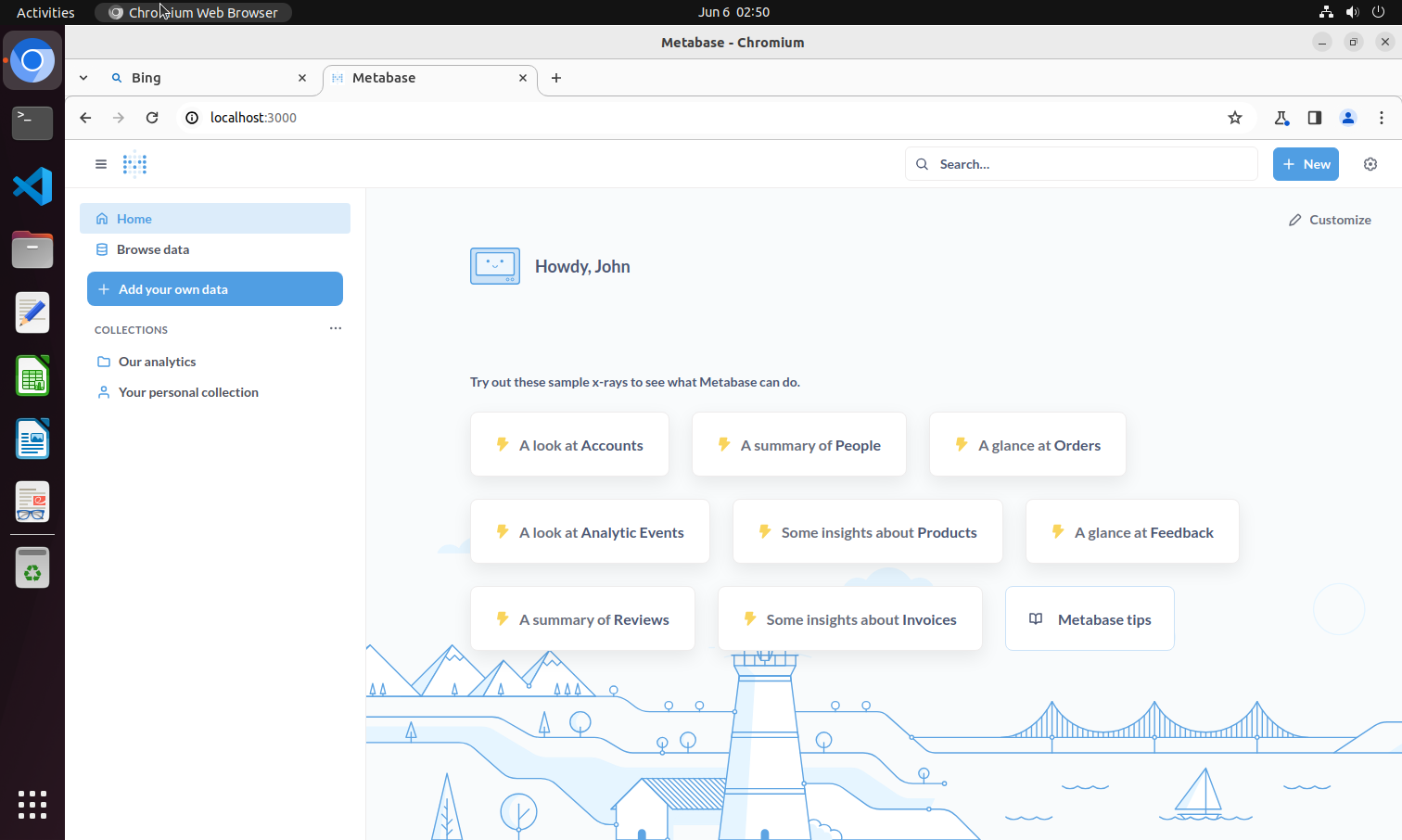}  \\\hline
Jupyter
Chromium & \textit{I want to use Logistic Regression to predict whether a student will be admitted to a college or not, and have now built the code framework in this open jupyter notebook. Please read the framework code and complete all the \#TODO sections. Finally, you need to run the code and save it.} & \includegraphics[width=6cm, height=3.38cm]{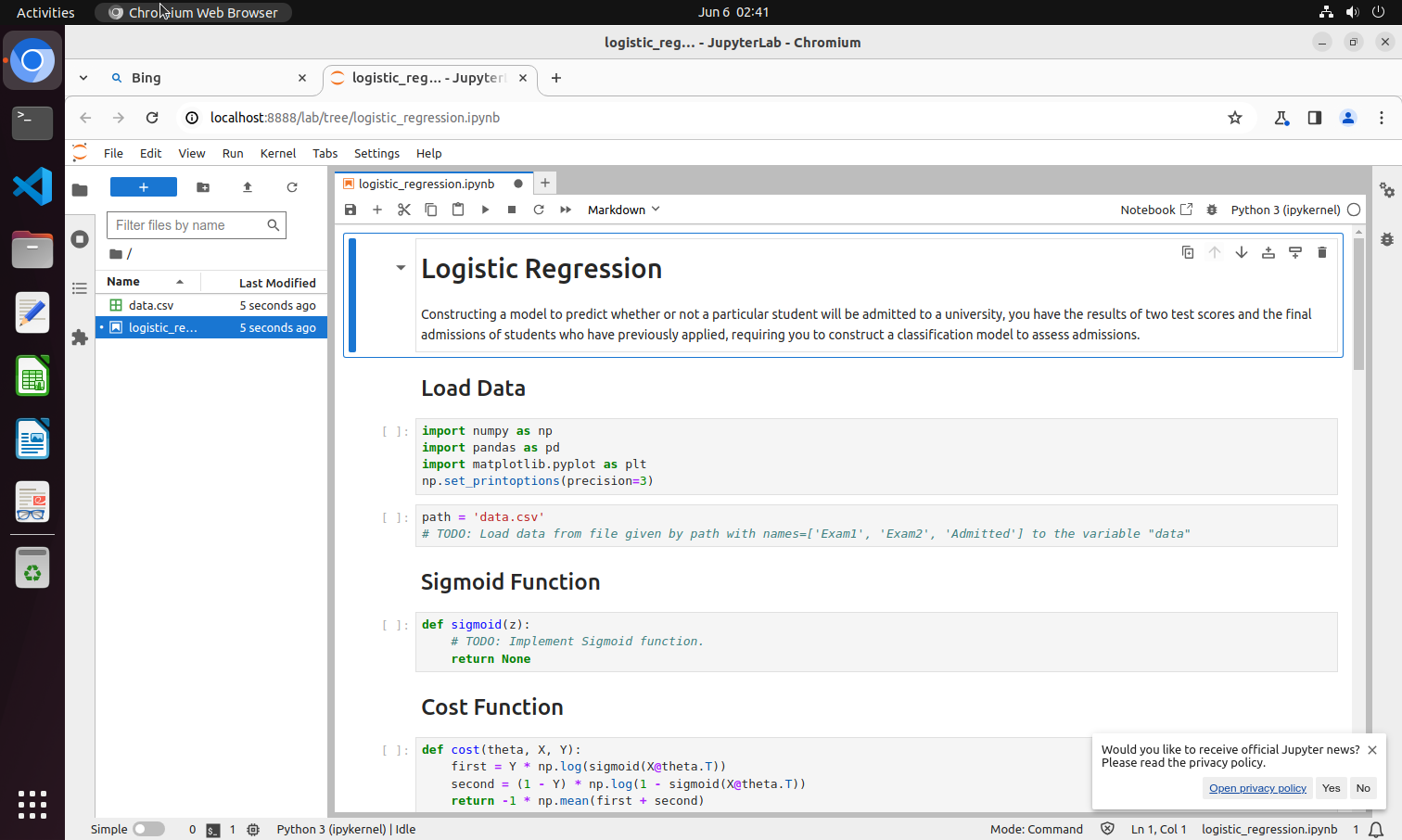} \\\hline
Excel & \textit{Add a new column named "Profit" and calculate the profit for each week by subtracting "COGS" from "Sales" in that column.} & \includegraphics[width=6cm, height=3.38cm]{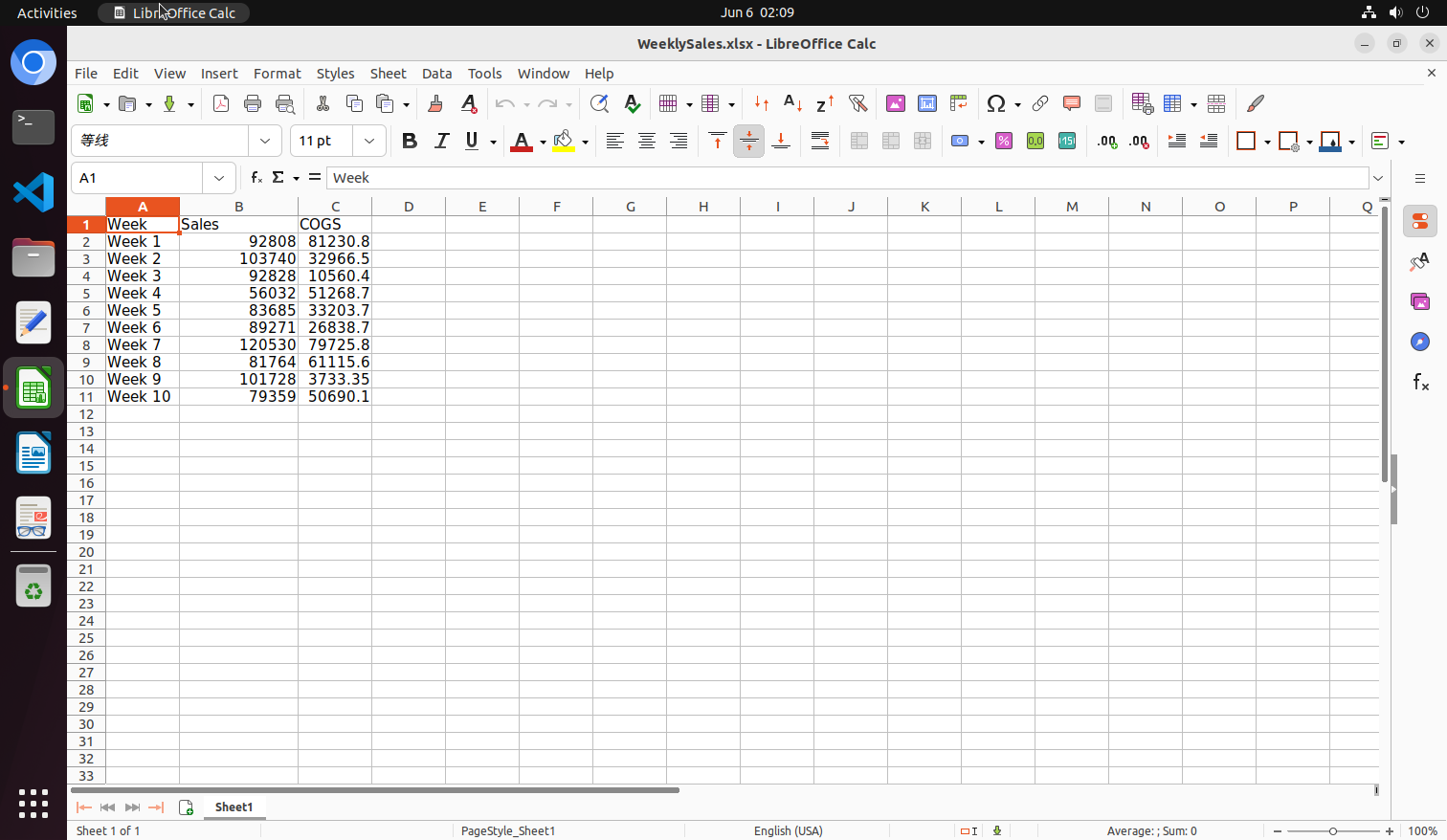} \\\hline
Superset
Chromium &  \textit{Help me create a rolling mean line chart for table flights to see the trend of the average cost per day. The rolling period should be 7 and save the chart as the name "rolling\_mean".} & \includegraphics[width=6cm, height=3.38cm]{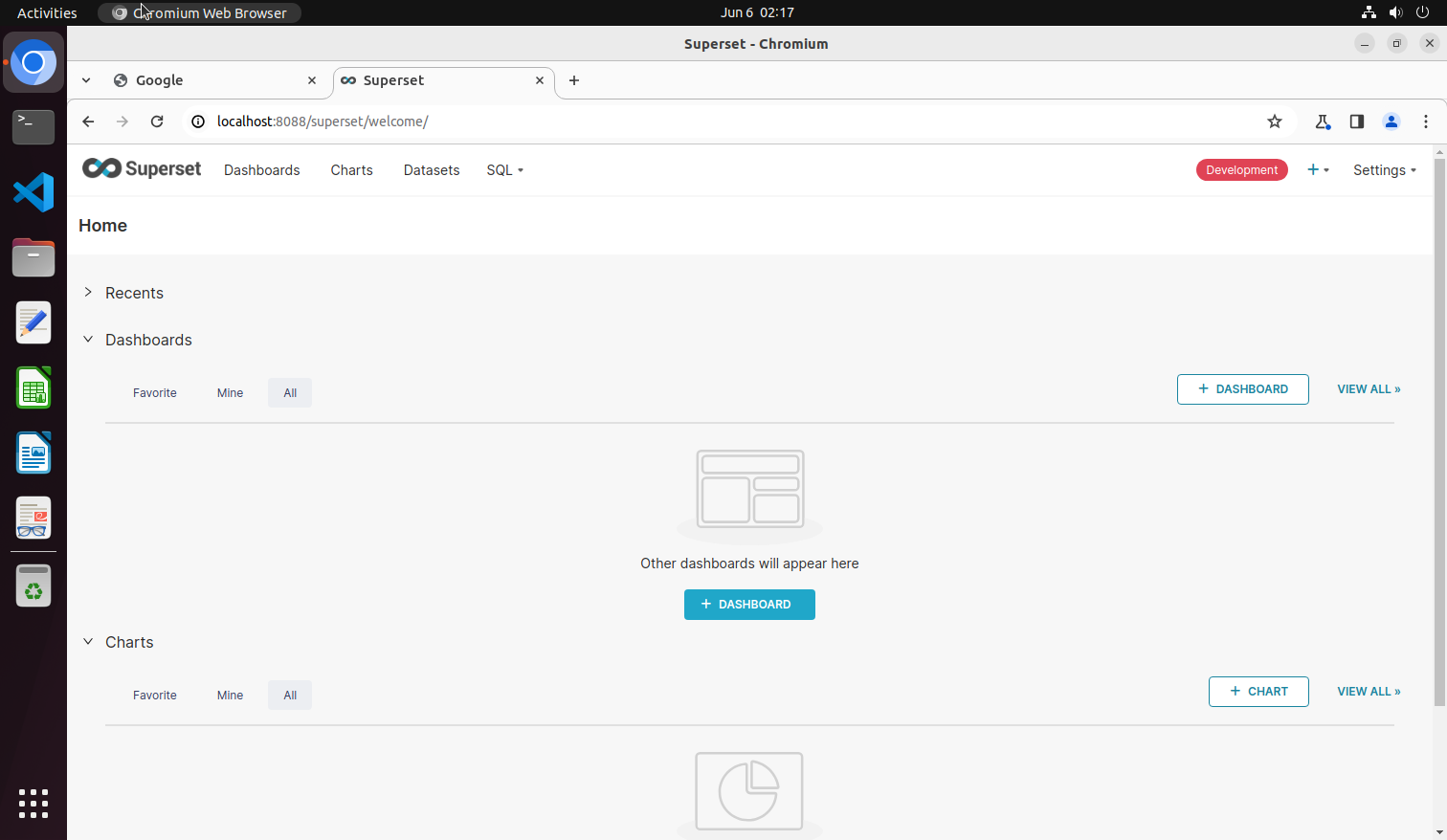} \\\hline
Airbyte
Chromium & \textit{Help me set up the destination of data transfer to a local JSON file in the Airbyte local UI. The target file path is /local/json\_destination.} & \includegraphics[width=6cm, height=3.38cm]{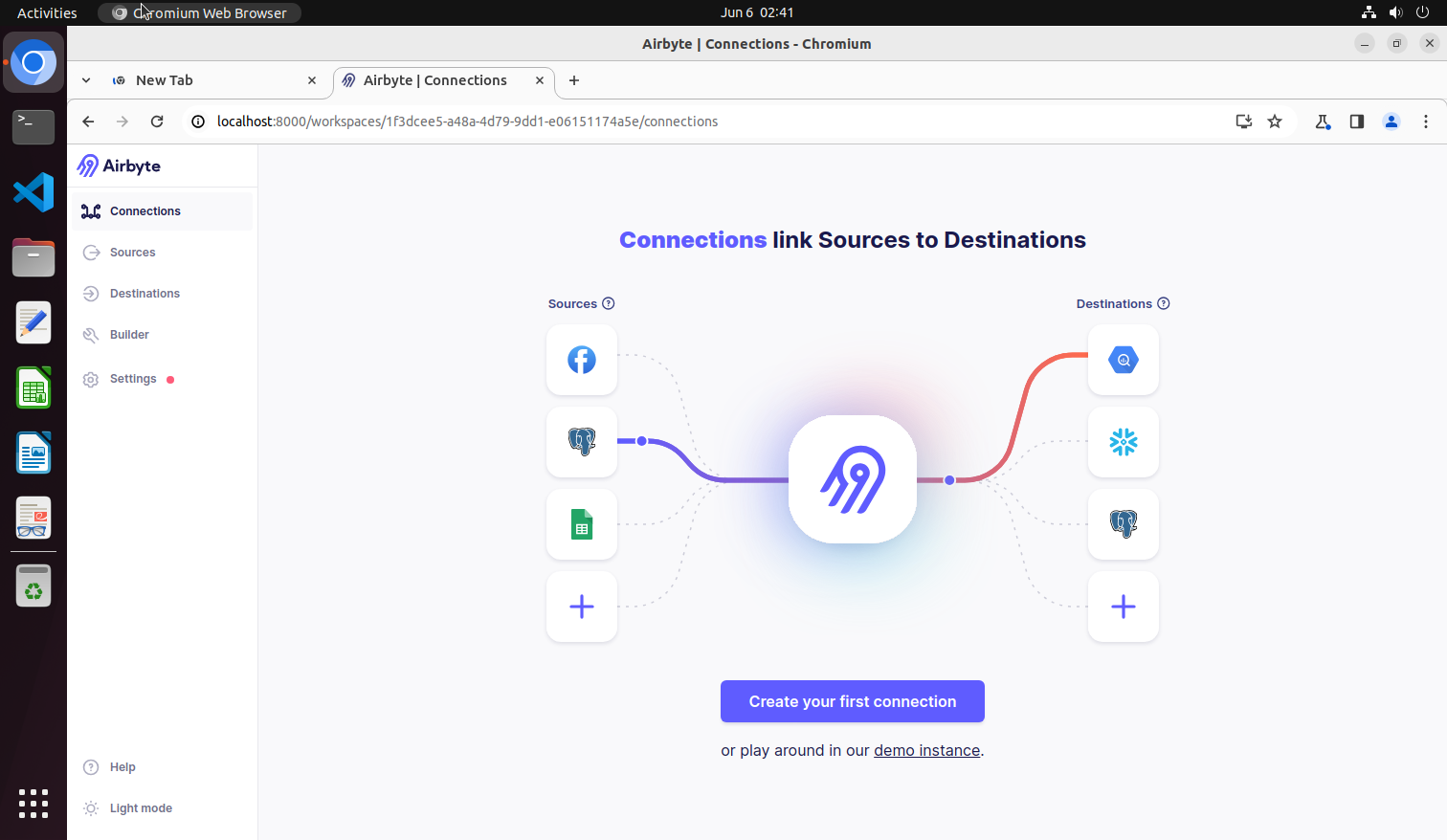} \\\hline
dbt-cloud
Chromium & \textit{I've created an empty dbt cloud project named "test\_connection". Could you help me set up the connection to a BigQuery GCP? You don't need to configure the repository for the project, and the credential file is provided at desktop.} & \includegraphics[width=6cm, height=3.38cm]{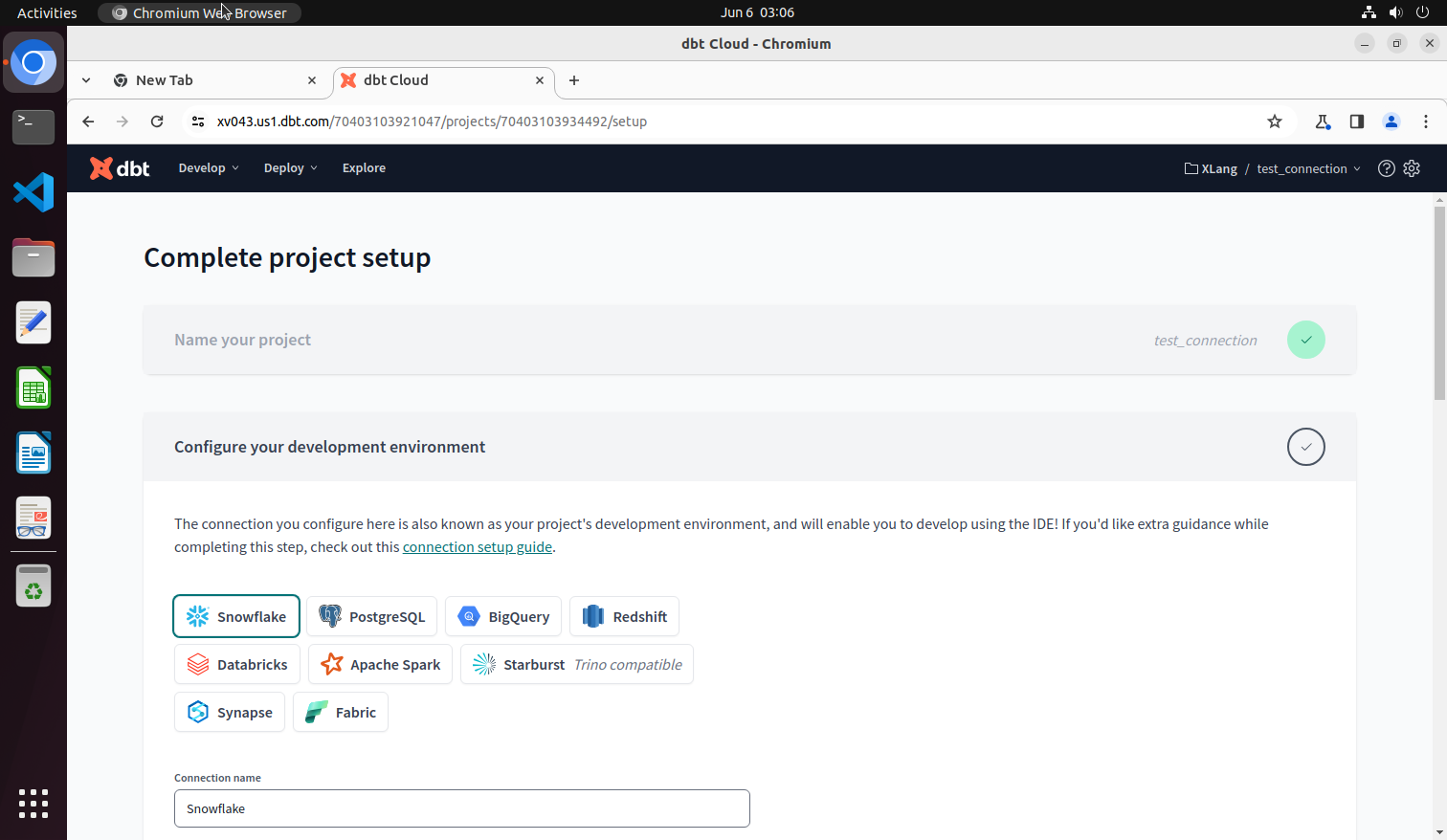} \\\hline
Airflow
Docker
VS Code\newline
Chromium & \textit{I have defined two DAGs to fetch and process data from TheCocktailDB. I hope to change the schedule of the consumer DAG such that each time the resulting files of the producer are updated, the consumer DAG is triggered. Can you help me with this data-aware scheduling?} & \includegraphics[width=6cm, height=3.38cm]{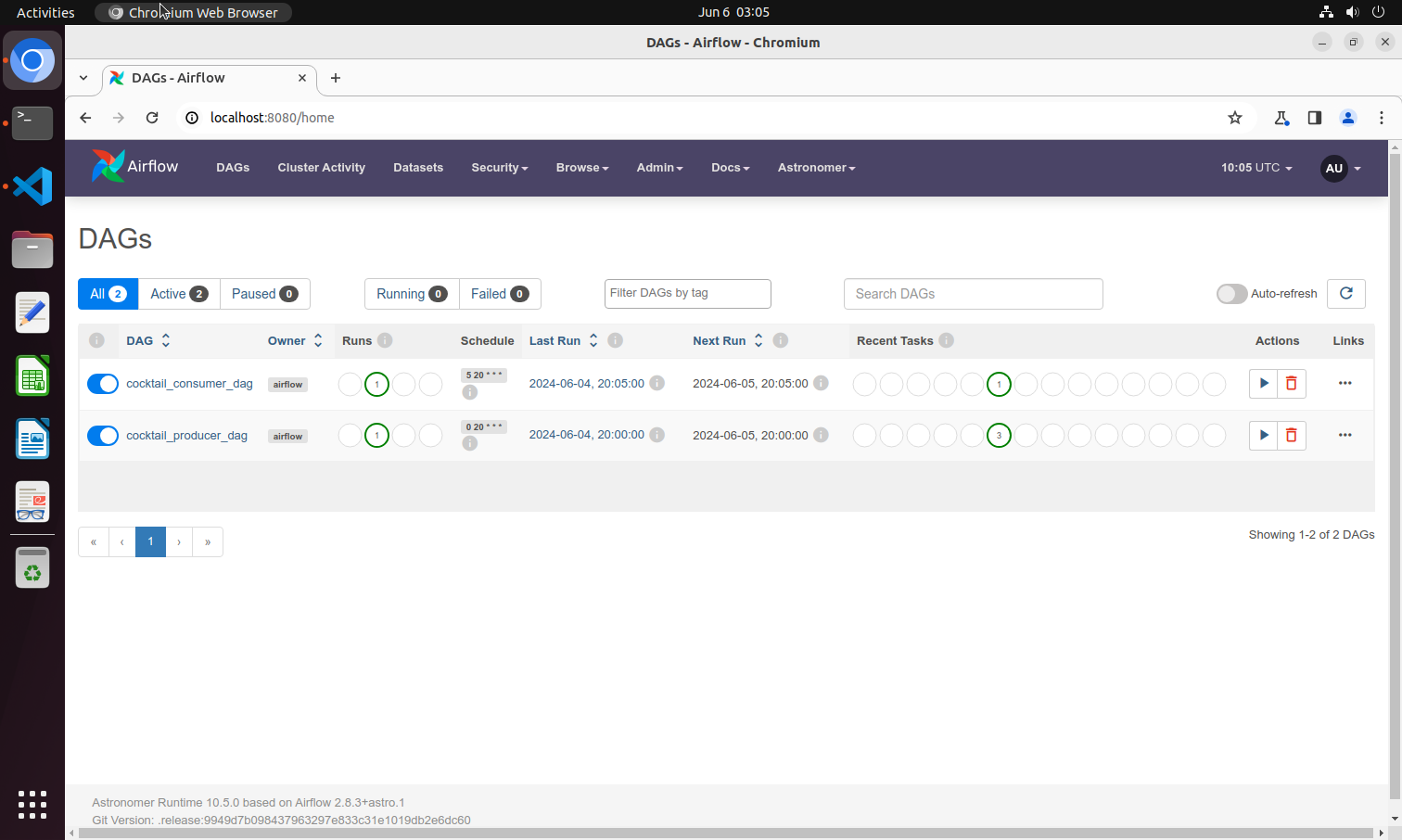} \\\hline
Dagster
Chromium
VS Code & \textit{Modify the current Dagster machine learning pipeline by adding two features "Age" and "Fare" to the Logistic Regression model from the data (you should fill in the NaN values by the mean of the column). Launch a run of the job "sklearn\_job", and schedule it to run at every hour on weekdays.} &\includegraphics[width=6cm, height=3.38cm]{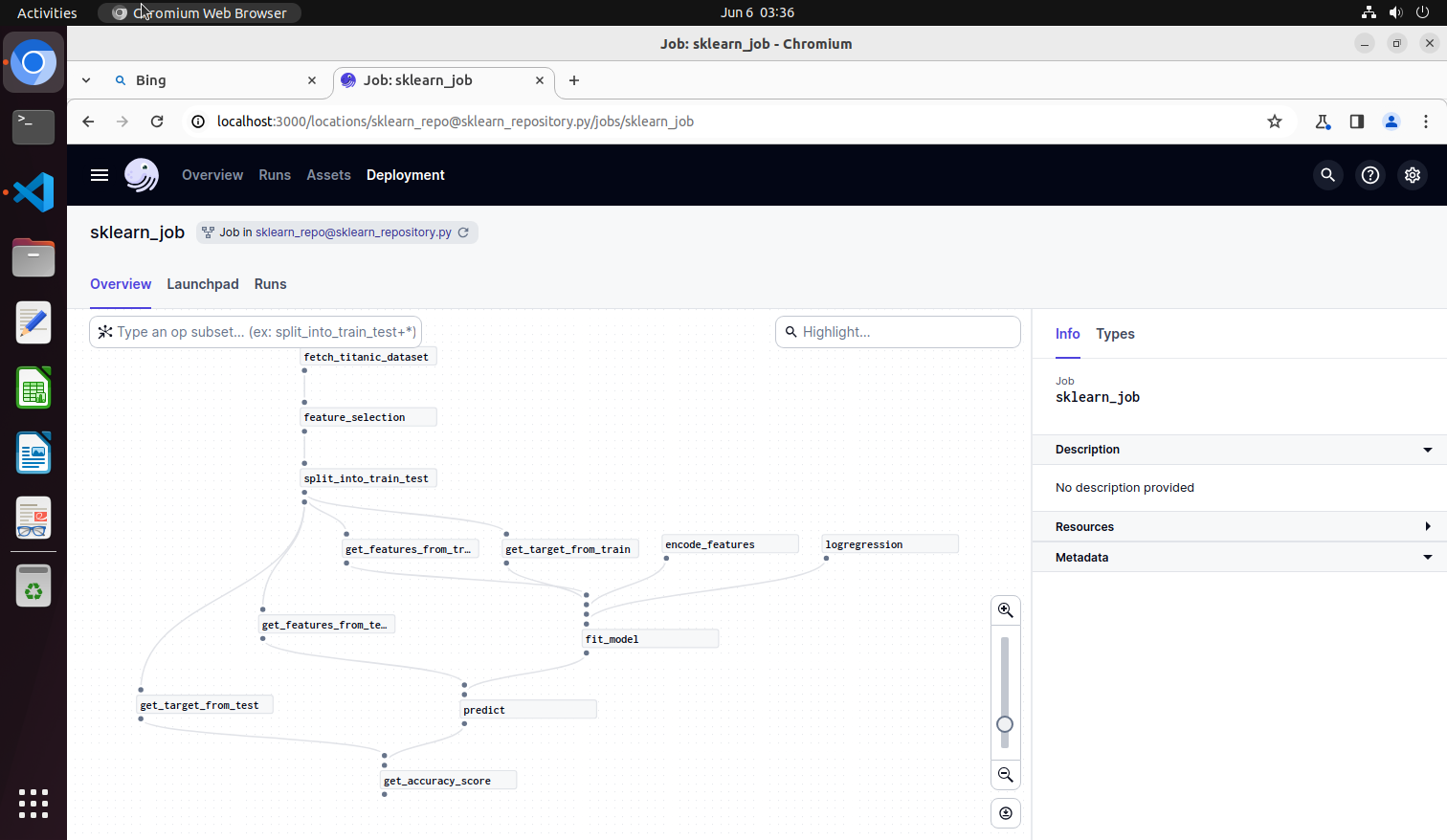} \\\hline
Snowflake
Chromium & \textit{I heard there are many free to download datasets on Snowflake marketplace. And I am really curious about worldwide addresses. Could you help me get one database about it? Name it `WORLDWIDE\_ADDRESSES`.} & \includegraphics[width=6cm, height=3.38cm]{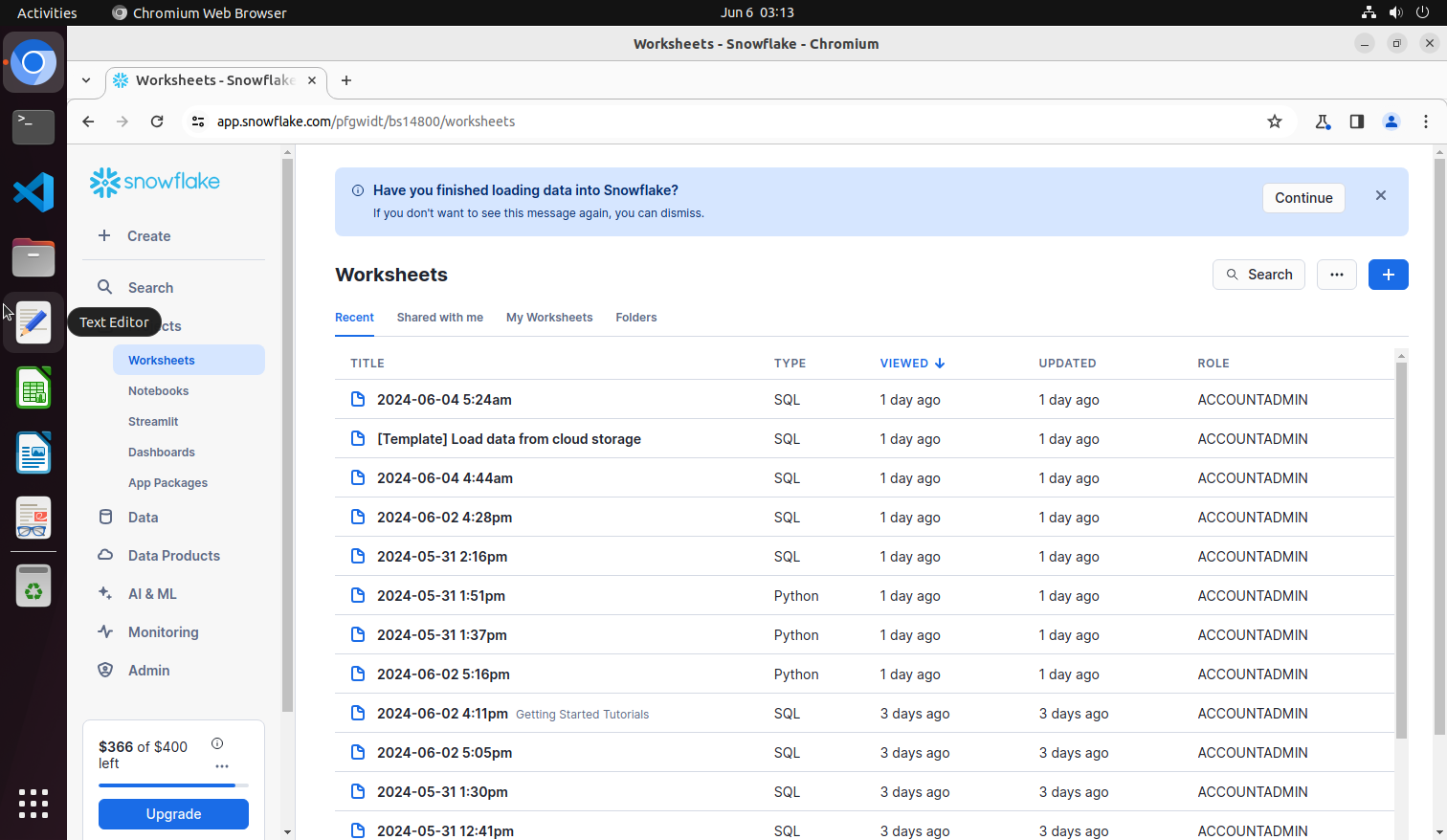} \\\hline
ServiceNow
Chromium & \textit{Go to the hardware store and order 8 "iPad mini" with configuration \{'Choose the colour': 'Purple', 'Choose the storage': '256'\}} & \includegraphics[width=6cm, height=3.38cm]{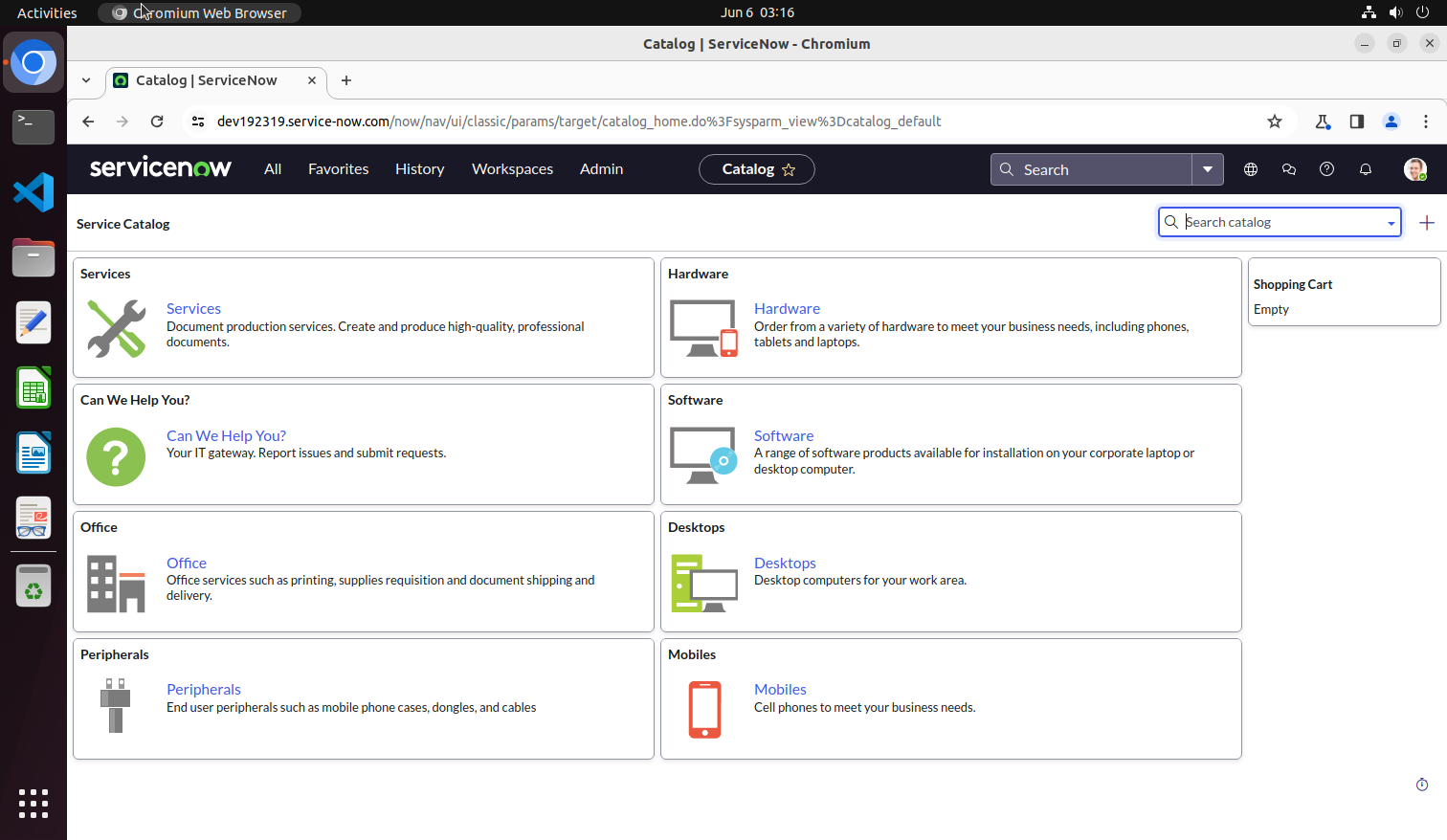}\\\hline
BigQuery
Chromium & \textit{Load the data from the Google drive Spider002 folder into Bigquery's 'data1' table of 'information' datasets.} & \includegraphics[width=6cm, height=3.38cm]{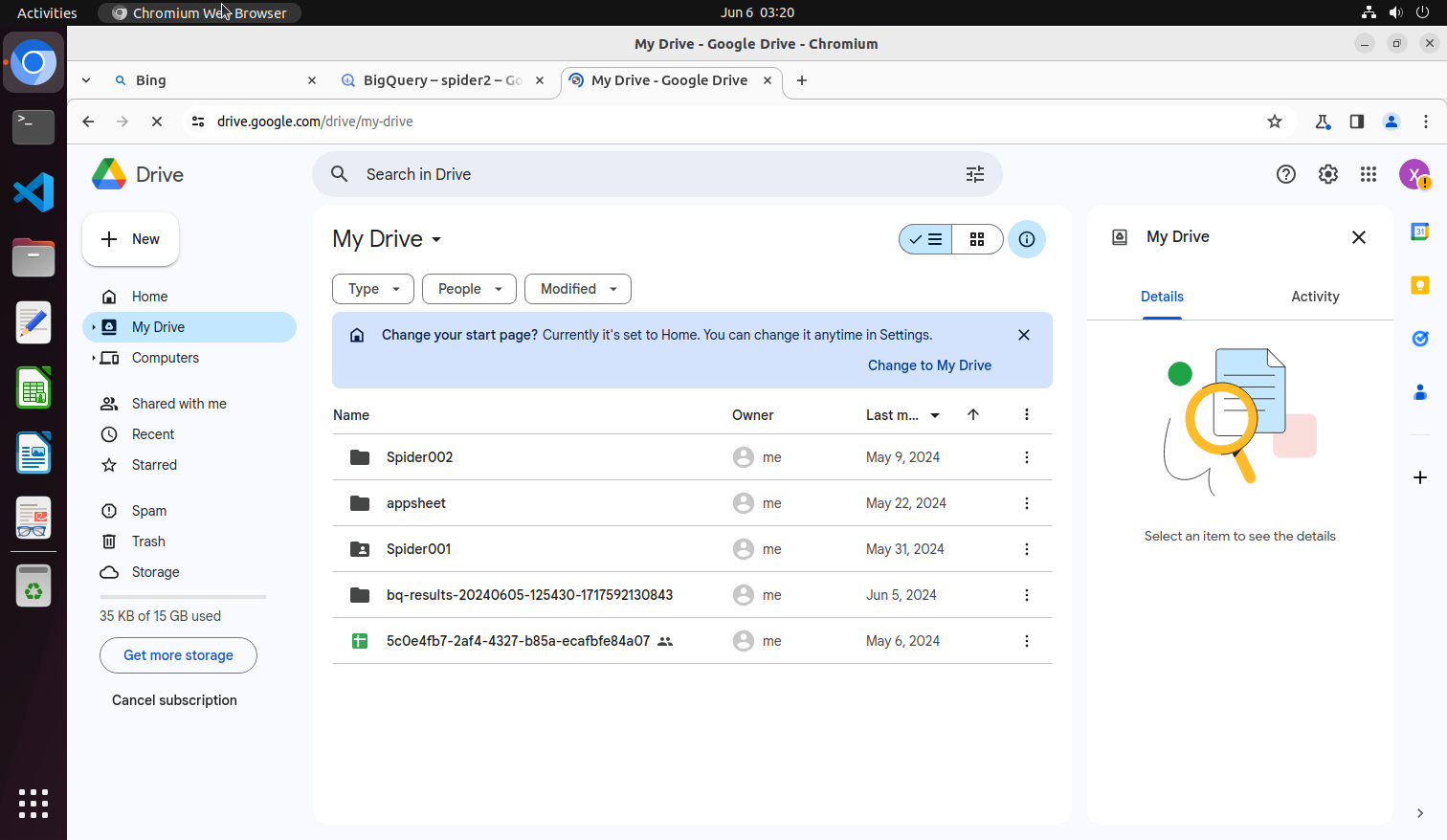} \\\hline
Metabase
Postgresql
Chromium & \textit{Help me finish the metabase login setup with information shown in setup.json.} & \includegraphics[width=6cm, height=3.38cm]{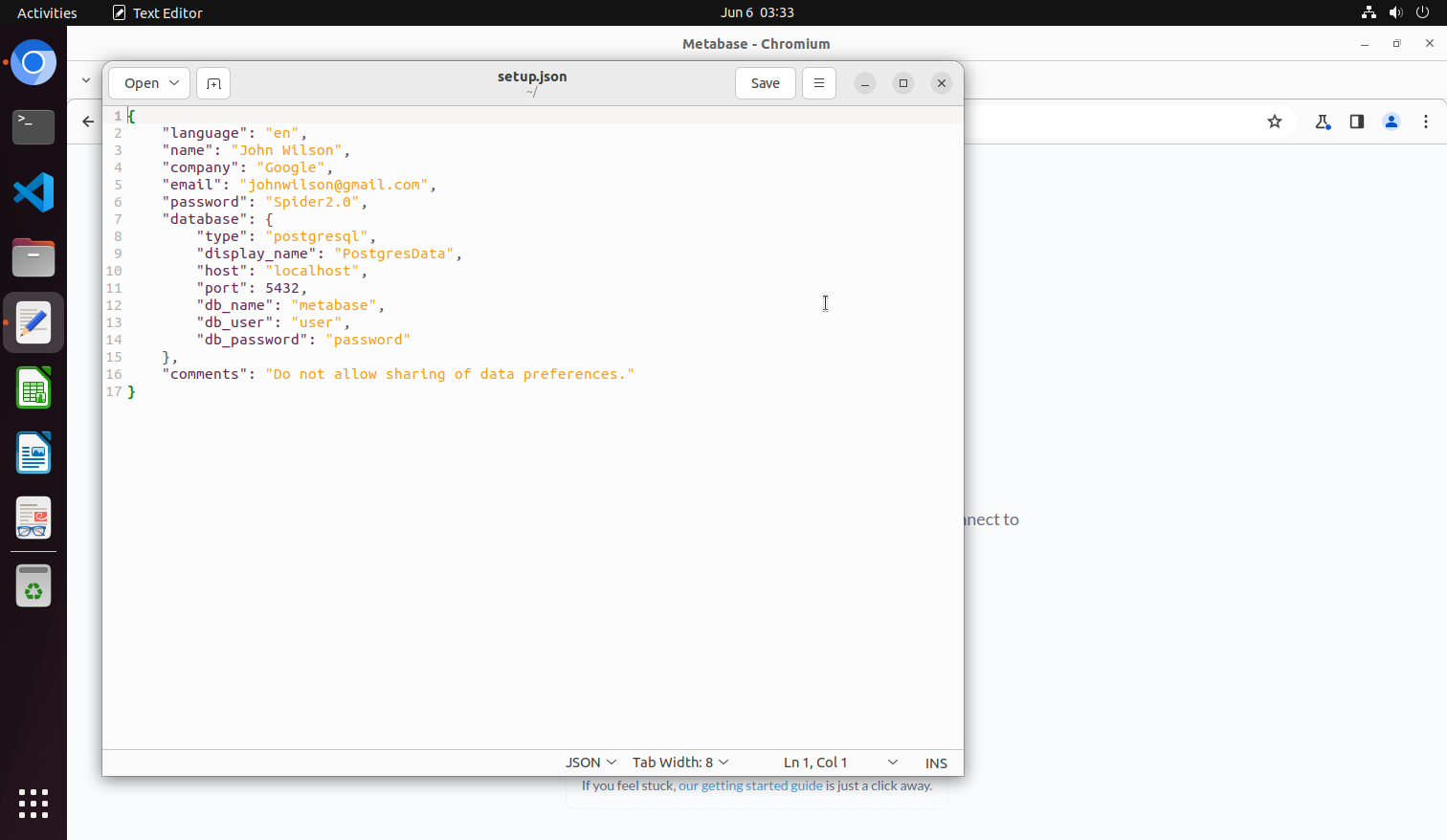} \\\hline
Dagster
Chromium
VS Code & \textit{I just built a 3-step Dagster pipeline. Now, I want to run it regularly to keep all assets up to date. Name the target job `hacker\_news\_pipeline` and schedule it to run every hour.} & \includegraphics[width=6cm, height=3.38cm]{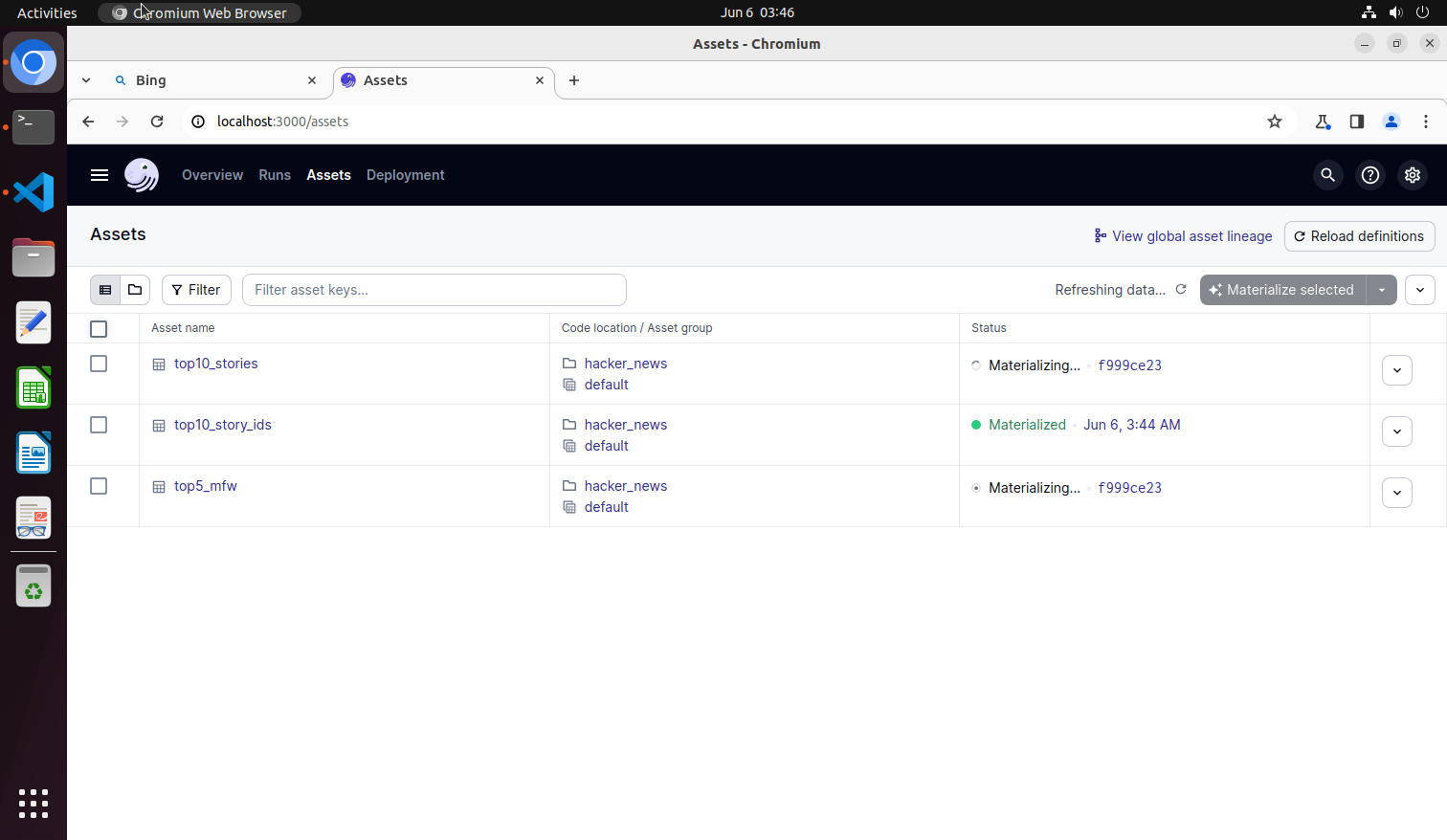} \\\hline
dbt-cloud
Chromium
Terminal & \textit{Install dbt-cloud-cli from GitHub and extract the binary to the same folder as the dbt project "analytics". Follow the instruction "Step 1: Install" specified in the opened account profile page.} & \includegraphics[width=6cm, height=3.38cm]{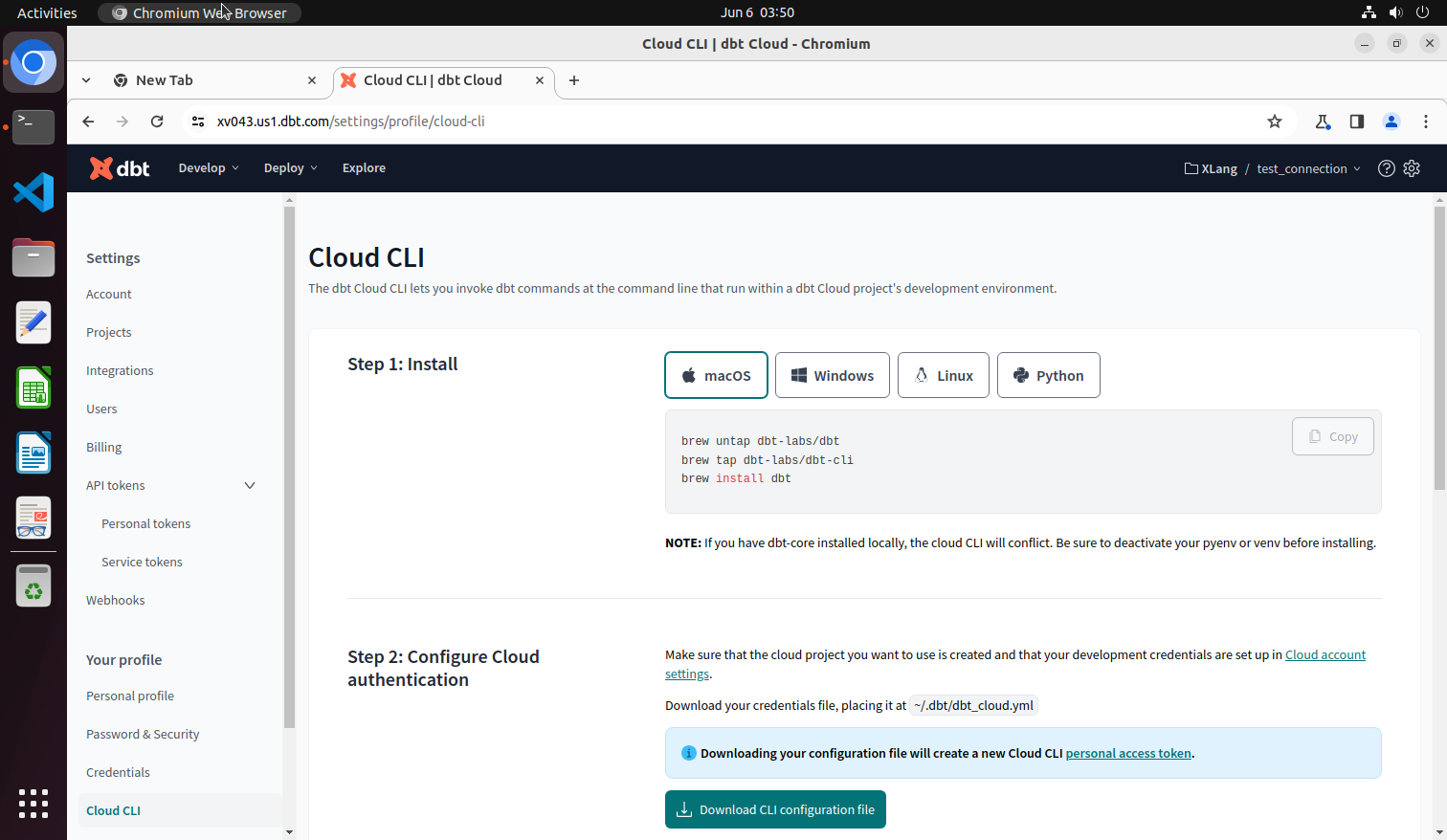} \\
\hline

\hline
\end{longtable}

%% file: appendices/prompts.tex
\clearpage

\section{Prompts for Multi-modal Agents}
\label{app:prompts}
Multi-modal agent baseline involves complex prompt engineering. The following sections will introduce the system prompt, task prompt, and retrieved context augmented prompt. 

\subsection{System Prompt}
The entire system prompt consists of the environment prompt, observation space prompt, action space prompt, and general tips. Different action/observation types have different prompts. In this section, we will introduce each one in turn and present the overall system prompt at last.

\subsubsection{Observation Space Prompt}
The four different observation space settings, namely 1) screenshot, 2) a11ytree, 3) screenshot+a11ytree, and 4) SoM, each has a different prompt.

\paragraph{Screenshot Setting}\mbox{}

\begin{tcolorbox}
\begin{Verbatim}[breaklines=true]
After each action step, you will get an image-style observation, which is the screenshot of the computer screen. And you need to predict the next action on the computer based on this image.
\end{Verbatim}
\end{tcolorbox}

\paragraph{Accessibility Tree Setting}\mbox{}

\begin{tcolorbox}
\begin{Verbatim}[breaklines=true]
After each action step, you will get a text-style observation, which is extracted and pruned from the accessibility tree based on AT-SPI library. The accessibility tree describes the elements (e.g., panels, icons, buttons, frames, windows, applications) on the computer desktop, as well as its embedded text content, status and positions. For simplicity, we prune the original tree and only extract useful information into a tabular format for you. Here is a quick glance on the observation:
TAG, NAME, POSITION (top-left x & y), SIZE (width & height), TEXT
menu, Visual Studio Code, (99, 0), (184, 27), ''
push-button, Chromium Web Browser, (0, 33), (70, 64), ''
terminal, Terminal, (70, 74), (1430, 832), '(base) user@ubuntu:~/projects/$'

... more rows ...

, where `TAG` / `NAME` is the element type / name respectively. `POSITION` and `SIZE` together describe the square position of this element on the computer screen. For example, if you want to click one button, you can click any point in the square area defined by `POSITION` and `SIZE`. Assume that the position of this button is (100, 200), and the size is (40, 40), the CENTER of this button is (120, 220), which is calculated by x = 100 + 40 / 2 = 120, y = 200 + 40 / 2 = 220. `TEXT` refers to the text content embedded in the element, e.g., the bash terminal output or texts in an editable input box.

And you will predict the next action of the computer based on the accessibility tree.
\end{Verbatim}
\end{tcolorbox}

\clearpage
\paragraph{Screenshot + Accessibility Tree Setting}\mbox{}

\begin{tcolorbox}
\begin{Verbatim}[breaklines=true]
The observation space is a combination of two sources: 1) image-style screenshot of the desktop, and 2) text-style accessibility tree derived from AT-SPI library.

### Screenshot

After each action step, you will get a image-style observation, which is the screenshot of the computer screen. And you need to predict the next action on the computer based on this image. You can use this image to locate the elements on the screen or check the status of the computer, especially whether the previous action is successful or not.

### Accessibility Tree

The accessibility tree describes the elements (e.g., panels, icons, buttons, frames, windows, applications) on the computer desktop, as well as its embedded text content, status and positions. For simplicity, we prune the original tree and only extract useful information into a tabular format for you. Here is a quick glance on the observation:

TAG, NAME, POSITION (top-left x & y), SIZE (width & height), TEXT
menu, Visual Studio Code, (99, 0), (184, 27), ''
push-button, Chromium Web Browser, (0, 33), (70, 64), ''
terminal, Terminal, (70, 74), (1430, 832), '(base) user@ubuntu:~/projects/$'

... more rows ...

, where `TAG` / `NAME` is the element type / name respectively. `POSITION` and `SIZE` together describe the square position of this element on the computer screen. For example, if you want to click one button, you can click any point in the square area defined by `POSITION` and `SIZE`. Assume that the position of this button is (100, 200), and the size is (40, 40), the CENTER of this button is (120, 220), which is calculated by x = 100 + 40 / 2 = 120, y = 200 + 40 / 2 = 220. `TEXT` refers to the text content embedded in the element, e.g., the bash terminal output or texts in an editable input box.

You can use the accessibility tree to accurately locate positions of useful elements on the screen and check the concrete textual contents of elements. 

By combining the screenshot and accessibility tree, you should be intelligent to predict the next feasible and meaningful action.
\end{Verbatim}
\end{tcolorbox}

\clearpage
\paragraph{SoM Setting}\mbox{}
\begin{tcolorbox}
\begin{Verbatim}[breaklines=true]
The observation space is a combination of two sources: 1) image-style screenshot of the desktop with interact-able elements marked with numerical indexes, and 2) text-style accessibility tree derived from AT-SPI library.

### Labeled Screenshot

After each action step, you will get a image-style observation, which is the screenshot of the computer screen. For ease of locating positions of elements, we extend the original screenshot with index marks. That is, some salient elements which can be interacted with (e.g., a button or editable input box) are marked with line boudaries and numeric indexes. You can use this image to locate the elements on the screen or check the status of the computer, especially whether the previous action is successful or not.

### Accessibility Tree

The accessibility tree describes the elements (e.g., panels, icons, buttons, frames, windows, applications) on the computer desktop, as well as its embedded text content, status and positions. For simplicity, we prune the original tree and only extract useful information into a tabular format for you. Here is a quick glance on the observation:
INDEX, TAG, NAME, POSITION(top-left x & y), SIZE(width & height),TEXT
1,   menu,        Visual Studio Code, (99, 0),  (184, 27),  ''
2,   push-button, Chromium Web Browser, (0, 33),    (70, 64),  ''
3,    terminal,   Terminal,           (70, 74),  (1430, 832), (base)
user@ubuntu:~/projects/$'
... more rows ... 

, where `INDEX` indicates exactly the numeric label for each element marked in the screenshot. You can use this alignment information to simplify your predicted action. For example, you can use `pyautogui.click(index_2)` to represent clicking the CENTER of the element with index 2 on the screenshot. We will automatically perform the position calculation and substitution for you. `TAG` / `NAME` is the element type / name respectively. `POSITION` and `SIZE` together describe the square position of this element on the computer screen. For example, if you want to click one button, you can click any point in the square area defined by `POSITION` and `SIZE`. Assume that the position of this button is (100, 200), and the size is (40, 40), the CENTER of this button is (120, 220), which is calculated by x = 100 + 40 / 2 = 120, y = 200 + 40 / 2 = 220. `TEXT` refers to the text content embedded in the element, e.g., the bash terminal output or texts in an editable input box.
You can use the accessibility tree to accurately locate positions of useful elements on the screen and check the concrete textual contents of elements. 
By combining the screenshot and accessibility tree, you should be intelligent to predict the next feasible and meaningful action.
\end{Verbatim}
\end{tcolorbox}

\clearpage
\subsubsection{Action Space Prompt}
\label{app:action_space_prompt}
As for the prompt of action space, we provide two choices: 1) pyautogui code, and 2) JSON dict.
\paragraph{pyautogui Code}\mbox{}
\begin{tcolorbox}
\begin{Verbatim}[breaklines=true]
You are required to use `pyautogui` to perform the action grounded to the observation. And the action space includes two types:

1. Python code block using pyautogui wrapped by 3 backticks, e.g.,
```python
# you python code here, e.g.,
pyautogui.hotkey('ctrl', 'c')
```

2. Three pre-defined special actions: [WAIT, FAIL, DONE]
- When you think you have to wait for some time, return ```WAIT```;
- When you think the task can not be done, return ```FAIL```, don't easily say ```FAIL```, try your best to do the task;
- When you think the task is done, return ```DONE```.
These 3 actions also need to be wrapped by 3 backticks.

### REMEMBER THAT:

0. We will import libraries `pyautogui` and `time` automatically for you, but if you use other python libraries, PLEASE IMPORT THEM FIRST ALTHOUGH THIS IS DISCOURAGED;
1. DONOT use the `pyautogui.locateCenterOnScreen` function to locate the element you want to operate with, since we have no image of the element you want to operate with;
2. DONOT use the `pyautogui.screenshot` function to make screenshot;
3. For time efficiency, you can return one line or multiple lines of python code to perform continuous actions in one response. For example, your response may contain the following code block:
```
pyautogui.moveTo(100, 210)
pyautogui.dragTo(500, 200, button='left', mouseDownUp=True)
pyautogui.rightClick()
```
4. When predicting multiple lines of code, make some small delay like `time.sleep(0.5)` interval, such that the machine can response correctly. And it is STRONGLY RECOMMENDED that, for one action which may influence the environment significantly (e.g., click the button of one application to open it, or click a web link which navigates to a new page), it is better to predict this action without follow-ups in order to observe the changes in environment states first;
5. Each time when you predict code, neither variables nor function is shared acrossed different code blocks. In other words, each code block will be executed in isolation;
6. For coordinates (x, y), please speculate or calculate by yourself based on the observation of previous interaction turn. BE CAREFUL to ensure the coordinates are feasible.
7. Please pay attention that, code wrapped by 3 backticks ``` will be recognized as an action in the action space. Therefore, when you output non-action code, please use other symbols like ''' instead.
\end{Verbatim}
\end{tcolorbox}

\clearpage
\paragraph{JSON Dict~(truncated)}\mbox{}
\begin{tcolorbox}
    \begin{Verbatim}[breaklines=true]
Firstly, we use json dict to describe the types and parameters for each action we allowed (`required=true` means this argument must be provided). Then, we demonstrate use cases, and precautions.

### Specification for All Actions

ACTION_LIST = [
    {
        "action_type": "MOVE_TO",
        "note": "move the cursor to a specified position (x, y)",
        "parameters": {
            "x": {
                "type": float,
                "range": [0, MAX_SCREEN_WIDTH],
                "required": true,
            },
            "y": {
                "type": float,
                "range": [0, MAX_SCREEN_HEIGHT],
                "required": true,
            }
        }
    },
    ... more action dicts ...
]

### Use Cases

- For MOVE_TO, you need to predict the x and y coordinate of the mouse cursor, the left top corner of the screen is (0, 0).
Use case: move the mouse to position (56.1, 65.0)
```json
{
    "action_type": "MOVE_TO",
    "x": 56.1,
    "y": 65.0
}

... more use cases ...

### Precautions

1) The output action MUST BE CHOSEN and CAN ONLY BE CHOSEN from the action space (json dict) defined above, otherwise your action will be considered as invalid and you will get a penalty. For example, bash, sql, or python code WILL NOT be executed;
2) For each action dict, STRICTLY OBEY THE FORMAT, which must contain the `action_type` field and required parameters. Optional parameters will be set to default values if not provided. NEVER RETURN ME ANYTHING ELSE WHICH IS NOT DEFINED;
3) For efficiency, you CAN predict multiple actions in one response, but REMEMBER TO WRAP EACH ACTION DICT SEPARATELY using backticks ```json and ```.
    \end{Verbatim}
\end{tcolorbox}

\clearpage
\subsubsection{Overall System Prompt}
\begin{tcolorbox}
    \begin{Verbatim}[breaklines=true]
You are an intellignet agent who is expert in completing data
science/engineering tasks using professional tools on computer. You have deep understanding of computer basics and data science/engineering knowledge.
Now, you will interact with a real desktop environment, which is an Ubuntu operating system that has access to the Internet. You should strictly follow the user instruction, communicate with the environment and try your best to complete the given data-related task successfully. Generally, you will communicate with the environment in this interactive and continuous manner:
1) In each iteration, you should take one action to control the keyboard or mouse in the desktop environment given the actions and observations from a few previous steps;
2) Then, you will obtain new observations from the environment after the action is grounded (you do not need to worry about the execution, we will perform it for you);
3) Repeat steps 1) and 2) until you think the work is done.

Here are the details of the action spaces (including usage and precautions) and observation spaces:

{{action_prompt}}

{{observation_prompt}}

Besides, here are some important tips for you to better complete the task:
1. My computer's password is 'password', feel free to use it when you need sudo rights.
2. The screen size for the running desktop is: ({screen_width}, {screen_height}).
3. Some action may need time to reflect in the environment (e.g., code execution and web page loading), please be patient and refer to the WAIT action.
4. Try to complete the task in as few steps as possible, we are on a tight budget.
5. Try to use the applications we opened for you as possible, e.g., use the opened gnome-terminal instead of the embedded one in Visual Studio Code.
6. For critical actions (e.g., opening an application or clicking a button), ensure the action succeeds before predicting or proceeding to the next one. That is, DO NOT be greedy to predict all actions all at once in one response without confirming the observation of a significant action.
7. When you try to write codes or texts, please ensure you have focused on the right window or input panel. If the input panel already has some texts, be careful that you may need to clear or selecting them before overwritting.
8. DO NOT be stubborn to complete the task in one step. You can break down the task into several steps and complete them one by one.
9. DO NOT be stupid to repeat the same actions without any progress. If you find that the action is not effective in the observation, try another one.
10. RETURN ME ONLY THE ACTION DEFINED IN ACTION SPACES. NEVER EVER RETURN ME ANYTHING ELSE. THIS IS CRITICAL!!!
    \end{Verbatim}
\end{tcolorbox}

\clearpage
\subsection{Task Prompt}
The task instruction for \ours has two forms. The abstract instruction describes the overall goal of a task without a step-by-step solution, thus testing both planning and grounding abilities. The verbose instruction provides a detailed tutorial-like solution to the task, primarily validating the grounding ability.

\subsubsection{Example of Task Prompt for Abstract Instructions}

\begin{tcolorbox}
\begin{Verbatim}[breaklines=true]
Now, let's start the task!
You are asked to complete the following task: I want to build an airflow project connecting to a local postgres database. Could you install docker, astro and postgresql for me. The sudo password is 'password' (' not included). By the way, configure docker and postgresql to auto-start on boot, and allow me to prevent typing sudo when using docker each time.
\end{Verbatim}
\end{tcolorbox}

\subsubsection{Example of Task Prompt for Verbose Instructions}

\begin{tcolorbox}
\begin{Verbatim}[breaklines=true]
Here is a step-by-step tutorial from an expert instructing you how to complete it:

Now we want to upload data from xlang_gcs/google_ads/ in google cloud storage to my dataset google_ads. To do this:
1. Click the "+ ADD" button next to the "Explorer" panel.
2. Click the "Google Cloud Storage" panel on the pop-up window.
3. In the input box "Google Cloud Storage", enter the 'xlang_gcs/google_ads/account_history_data.csv' in the second windows. This window is labeled 'Select file from GCS bucket or use a a URI pattern'.
4. Destination Part, set Dataset to 'my_google_ads'
5. In Destination Part, set Table to 'account_history_data'
6. In Schema part, Mark the check mark in front of Auto detect.
7. Then, click the blue `CREATE TABLE` button at the bottom.
8. After page loading, click the "+ ADD" button next to the "Explorer" panel again.
9. Click the "Google Cloud Storage" panel on the pop-up window.
10. In the input box "Google Cloud Storage", enter the 'xlang_gcs/google_ads/account_stats_data.csv' in the second windows. This window is labeled 'Select file from GCS bucket or use a a URI pattern'.
11. Destination Part, set Dataset to 'my_google_ads'
12. In Destination Part, set Table to 'account_stats_data'
13. In Schema part, Mark the check mark in front of Auto detect.
14. Click the `CREATE TABLE` button at the bottom left in the pop-up window.
Eventually, we have completed this task.

You can exactly follow the detailed plan above or proactively tackle the task based on the real-time environment interaction by yourself.
\end{Verbatim}
\end{tcolorbox}

\clearpage
\subsection{Example of Retrieved Context Augmented Task Prompt}
\label{app:rag_prompt}
We also introduce a RAG setting, where we collect and clean the official documents of the professional tools as the retrieval corpus. We select top $k$~($k$ may depend on the constraint on input length) chunks~(each chunk is a token sequence with maximum length $512$) and insert them into the prompt input. Here are three demonstrations of different formats of the retrieved context.
\paragraph{Pure Text Format}\mbox{}
\begin{tcolorbox}
\begin{Verbatim}[breaklines=true]
We also retrieve relevant documentation from the web to help you with the task:

Documentation Source:
release-1-7-2.dagster.dagster-docs.io/integrations/dagstermill/using- notebooks-with-dagster.html

Documentation Title:
Using Jupyter notebooks with Papermill and Dagster Tutorial

Documentation Content:
The page will display the notebook asset in the Asset Graph.
If you click the notebook asset, a sidebar containing info about the asset will slide out from the right side of the page. In the Description
section of the panel is a View Source Notebook button:
This button allows you to view the notebook directly in the UI. When clicked, Dagster will render the notebook - referenced in the
notebook_path parameter - that'll be executed when the iris_kmeans_jupyter asset is materialized:
Click the Materialize button. To view the execution as it happens, click the View button in the alert that displays.
After the run completes successfully, you can view the executed notebook in the UI. Click the asset again and locate the View Notebook button in the Materialization in Last Run section of the sidebar:
Click the button to display the executed notebook - specifically, the notebook that was executed and written to a persistent location:
Step 5: Add an upstream asset #
While our iris-kmeans notebook asset now materializes successfully, there are still some improvements we can make. The beginning of the notebook fetches the Iris dataset, which means that every time the notebook is materialized, the data is re-fetched.
To address this, we can factor the Iris dataset into its own asset. This will allow us to:
Use the asset as input to additional notebooks.
This means all notebooks analyzing the Iris dataset will use the same source data, which we only have to fetch once.
Materialize notebooks without fetching data for each materialization.
Instead of making potentially expensive API calls, Dagster can fetch the data from the previous materialization of the Iris dataset and provide that data as input to the notebook.
\end{Verbatim}
\end{tcolorbox}

\clearpage
\paragraph{Markdown Syntax Format}\mbox{}
\begin{tcolorbox}
\begin{Verbatim}[breaklines=true]
We also retrieve relevant documentation from the web to help you with the task:

Documentation Source:
release-1-7-2.dagster.dagster-docs.io/integrations/dagstermill/using-
notebooks-with-dagster.md

Documentation Title:
Using Jupyter notebooks with Papermill and Dagster Tutorial

Documentation Content:
When clicked, Dagster will render the notebook - referenced in the `notebook_path`parameter - that'll be executed when the `iris_kmeans_jupyter`asset is materialized:

!Click the **Materialize**button. To view the execution as it happens, click the **View**button in the alert that displays.

After the run completes successfully, you can view the executed notebook in the UI. Click the asset again and locate the **View Notebook**button in the **Materialization in Last Run**section of the sidebar:

!Click the button to display the executed notebook - specifically, the notebook that was executed and written to a persistent location:

!Step 5: Add an upstream asset#
------------------------------

While our `iris-kmeans`notebook asset now materializes successfully, there are still some improvements we can make. The beginning of the notebook fetches the Iris dataset, which means that every time the notebook is materialized, the data is re-fetched.

To address this, we can factor the Iris dataset into its own asset. This will allow us to:

**Use the asset as input to additional notebooks.**This means all notebooks analyzing the Iris dataset will use the same source data, which we only have to fetch once.

**Materialize notebooks without fetching data for each materialization.**Instead of making potentially expensive API calls, Dagster can fetch the data from the previous materialization of the Iris dataset and provide that data as input to the notebook.

In this step, you'll:

Create the Iris dataset assetProvide the Iris dataset as input to the notebookModify the notebook
\end{Verbatim}
\end{tcolorbox}

\clearpage
\paragraph{Simplified HTML Format}\mbox{}
\begin{tcolorbox}
\begin{Verbatim}[breaklines=true]
We also retrieve relevant documentation from the web to help you with the task:

Documentation Source:
release-1-7-2.dagster.dagster-docs.io/integrations/dagstermill/using-
notebooks-with-dagster.html

Documentation Title:
Using Jupyter notebooks with Papermill and Dagster Tutorial

Documentation Content:
If you execute these cells, several plots of the Iris dataset will be created:
<p>Next, we conduct our K-means analysis:</p>
<code>estimator <span>=</span>sklearn<span>.</span>cluster<span>.</span>KMeans
<span>(</span>n_clusters<span>=</span><span>3</span><span>)</span>
estimator<span>.</span>fit<span>(</span>iris<span>[</span>
<span>[</span><span>"Sepal length (cm)"</span><span>,</span>
<span>"Sepal width (cm)"</span><span>,</span>
<span>"Petal length (cm)"</span><span>,</span>
<span>"Petal width (cm)"</span>
<span>]</span><span>]</span><span>)</span>
</code>
<p>Lastly, we plot the results of the K-means analysis. From the plots, we can see that one species of Iris is separable from the other two, but a more sophisticated model will be required to distinguish the other two species:</p>
<p>Like many notebooks, this example does some fairly sophisticated work, including producing diagnostic plots and a statistical model. For now, this work is locked away in the <code>.ipynb</code>format, only reproducible using a complex Jupyter setup, and only programmatically accessible within the notebook context. We'll address this in the remainder of the tutorial.</p>
<h2>Step 2: Create a Dagster asset from the Jupyter Notebook<span>#</span></h2>
<p>By creating a Dagster asset from our notebook, we can integrate the notebook as part of our data platform. This enables us to make its contents more accessible to developers, stakeholders, and other assets in Dagster.</p>
<p>To create a Dagster asset from a Jupyter notebook, we can use the <code>define_dagstermill_asset</code>function.
\end{Verbatim}
\end{tcolorbox}